\documentclass[sigconf]{acmart}
\usepackage{algorithm}
\usepackage{algpseudocode}
\usepackage{blindtext}
\usepackage{todonotes}
\usepackage{multirow}
\usepackage{subcaption}
\usepackage{listings}
\usepackage{amsmath,amsfonts}
\usepackage{graphicx}
\usepackage{textcomp}
\usepackage{comment}
\usepackage{xcolor}
\usepackage{booktabs}
\usepackage{booktabs}
\usepackage{tikz}

\AtBeginDocument{%
  }

\newcommand{\circlenum}[1]{\tikz[baseline=(myanchor.base)] \node[circle,fill=.,inner sep=1pt] (myanchor) {\color{-.}\bfseries\footnotesize #1};}

\newcommand{\etal}{et al.~}
\newcommand{\ie}{i.e.,~}
\newcommand{\eg}{e.g.,~}

\long\def\authornote#1{%
	\leavevmode\unskip\raisebox{-3.5pt}{\rlap{$\scriptstyle\diamond$}}%
	\marginpar{\raggedright\hbadness=10000
		\def\baselinestretch{0.8}\tiny
		\it #1\par}}
\newcommand{\karthik}[1]{\authornote{KP: #1}}
\newcommand{\pritam}[1]{\authornote{PD: #1}}
\newcommand{\sysname}{{\em DeLorean}\xspace}
\newcommand{\mbr}{Model-based~}
\newcommand{\cbr}{Checkpoint-based~}
\newcommand{\lqr}{LQR-O~}

\setcopyright{acmcopyright} 
\copyrightyear{2024}
\acmYear{2024}
\acmDOI{XXXXXXX.XXXXXXX}

\acmConference[AsiaCCS'24]{ACM ASIA Conference on Computer and Communications Security}{July 01--05, 2024}{Singapore}

\acmPrice{15.00}
\acmISBN{978-1-4503-XXXX-X/18/06}

\begin{document}
\title{Diagnosis-guided Attack Recovery for Securing Robotic Vehicles from Sensor Deception Attacks}

\author{Pritam Dash}
\email{pdash@ece.ubc.ca}
\affiliation{
  \institution{University of British Columbia, Canada}
  \country{}
}
\author{Guanpeng Li}
\email{guanpeng-li@uiowa.edu}
\affiliation{
    \institution{University of Iowa, USA}
    \country{}
}
\author{Mehdi Karimibiuki}
\email{mkarimib@ece.ubc.ca}
\affiliation{
\institution{University of British Columbia, Canada}
  \country{}
}
\author{Karthik Pattabiraman}
\email{karthikp@ece.ubc.ca}
\affiliation{
\institution{University of British Columbia, Canada}
  \country{}
}

\renewcommand{\shortauthors}{Dash et al.}

\begin{abstract}
Sensors are crucial for perception and autonomous operation in robotic vehicles (RV). Unfortunately, RV sensors can be compromised by physical attacks such as sensor tampering or spoofing. 
In this paper, we present \sysname, a unified framework for attack detection, attack diagnosis, and recovering RVs from sensor deception attacks (SDA). 
\sysname can recover RVs even from strong SDAs in which the adversary targets multiple heterogeneous sensors simultaneously.
We propose a novel attack diagnosis technique that inspects the attack-induced errors under SDAs, and identifies the targeted sensors using causal analysis. 
\sysname then uses historic state information to selectively reconstruct physical states for compromised sensors, enabling targeted attack recovery under single or multi-sensor SDAs. 
We evaluate \sysname on four real and two simulated RVs under SDAs targeting various sensors, and we find that it successfully recovers RVs from SDAs in $93\%$ of the cases. 
\end{abstract}


\maketitle

\section{Introduction}
\label{sec:introduction}
Autonomous Robotic Vehicles (RV) such as drones and rovers 
RVs 
rely on their onboard sensors to perceive their physical states (e.g., position, angular orientation). 
Based on the physical states, specialized algorithms plan the RV's autonomous navigation in a feedback control loop~\cite{feedback-control}.
Unfortunately, attackers can feed erroneous sensor measurements to the RV through its physical channels (i.e., physical attacks), causing the RV to deviate from its course and/or crash~\cite{gpsspoofing1, gyroscopespoofing}. 
Physical attacks such as GPS spoofing have occurred in the wild on military drones~\cite{us-drone-gps-spoofing} and marine systems~\cite{uk-ship-gps-spoofing}.

We refer to physical attacks on RVs that target one or more sensors simultaneously as {\em Sensor Deception Attacks} (SDAs). SDAs 
have been demonstrated in autonomous driving systems~\cite{cam-lidar, sensor-confusion}, and can also be extended to RVs.  
SDAs cannot be mitigated by traditional security techniques as they target the sensor hardware rather than the software. 
Further, attack detection~\cite{savior, choi} is not enough, as the RV may 
crash after the detection~\cite{pid-piper}. 
Many drones have hardware fail-safe mechanisms (e.g., landing upon failure). However, activating these upon attack detection may not be safe, as the drone may land in adverse environments or fall into attackers' hands~\cite{reaper}. Therefore, it is important to recover RVs from SDAs. 

Prior work on attack recovery from physical attacks~\cite{recovery-lqr, recovery-lp, pcb-kong}, uses the RV's historic state information to derive safe control actions under the attack. 
However, these techniques assume that all the sensors in the RV are under attack (\ie a worst-case recovery scenario) regardless of the number of sensors actually attacked.
As a result, the recovery technique performs either overly conservative or overly aggressive recovery actions during the RV mission that leads to unnecessary mission disruptions and delays (shown later in the paper).
This is because, under the worst-case assumptions, prior recovery techniques isolate all the sensors and use historic states to derive recovery control actions. Thus, they lack real-time sensor feedback and accurate information about the RV's current state.
Attack recovery must ensure minimal deviations from the set path and ensure timely recovery without delaying critical tasks. 

Furthermore, attackers can launch stealthy attacks on RVs, causing small disruptions that build up over time~\cite{stealthy-attacks}. 
While techniques have been proposed to detect such stealthy attacks~\cite{pid-piper, savior}, they incur a delay in detection due to the controlled nature of these attacks. 
As attack recovery techniques~\cite{recovery-lqr, recovery-lp, pcb-kong} rely on the RV's historic states information, the detection delay degrades the effectiveness of attack recovery. 
This is because the RV may have significantly deviated from the desired path before the stealthy attack is flagged. 


The above issues arise from the lack of attack diagnosis, and a unified method to attack detection and recovery in RVs.
To secure RVs from SDAs, we need to: 
(i) reliably identify sensors targeted by the SDA and selectively remove them from the RV's feedback control loop, and 
(ii) bridge the disconnect between attack detection and recovery to prevent the impact of detection delay on recovery. 

In this paper, we propose \sysname, a unified framework for attack detection,  diagnosis, and recovery from SDAs in RVs. We propose a {\em graph-based probabilistic attack diagnosis technique} to identify the sensors targeted by the SDA (\ie attack diagnosis), and integrate the diagnosis with existing attack recovery techniques~\cite{recovery-lqr, recovery-lp}. 
This eliminates the worst-case assumptions in recovery techniques and enables targeted attack recovery.
Finally, we design methods to integrate existing attack detection~\cite{pid-piper, savior, choi} with recovery techniques to address the attack detection delay, which ensures safe recovery of the RV even from stealthy attacks. 

\sysname assumes there is an existing attack detection technique deployed on the RV~\cite{choi, savior, pid-piper}. 
After the attack detector raises an alert, \sysname activates attack diagnosis to reason about the abnormalities in the RV's physical states under an SDA, identifies the sensors targeted by the attack, and selectively removes those sensors from the RV's feedback control loop.
As RVs use sensor fusion to derive state estimates, it is challenging to attribute abnormalities in physical states to a specific sensor being under attack. 
We address this challenge by monitoring error inflation in \emph{all} physical state estimates derived from the RV's sensors, and perform causal analysis using factor graphs~\cite{factor-graph} to identify targeted sensors. 
The attack diagnosis is independent of the analysis performed by the attack detection techniques~\cite{choi, savior, pid-piper}. 
This helps mask false positives in the attack detection, 
as it is unlikely for both the detection and diagnosis techniques to incur false positives simultaneously. 


\sysname also introduces an automated method to record attack-free historic state information of the RV while accounting for the attack detection delay. 
Based on attack diagnosis outcomes, \sysname estimates the physical states corresponding to the compromised sensors using the historic state information, and incorporates the uncompromised sensor estimates to derive reliable RV state estimates despite the attack. 
Finally, the attack recovery technique uses these reliable state estimates to perform attack recovery. 



By unifying attack detection, diagnosis, and recovery, \sysname provides a targeted approach to recover RVs from both single and multi-sensor SDAs including stealthy attacks. 
{\em To the best of our knowledge, \sysname is the first attack diagnosis-guided attack recovery technique for RVs.}
Our contributions are as follows: 
\begin{itemize}
    \item Propose a graph-based probabilistic diagnosis approach to 
    identify the RV's sensors targeted by SDAs through causal analysis, independent of the attack detection 
    \item Propose automated methods to record attack-free historic state information, and state reconstruction strategies that enable existing attack recovery techniques~\cite{recovery-lp, recovery-lqr, pcb-kong} to perform targeted attack recovery under SDAs. 
    \item Design \sysname, a unified framework that integrates existing attack detection and attack recovery techniques with the proposed attack diagnosis technique, to provide diagnosis-guided attack recovery from SDAs. 
    \item Evaluate \sysname on six RVs - four real RVs and two simulated systems in a wide range of scenarios, 
    and under SDAs targeting different numbers of sensors in the RV. 
\end{itemize}

Our experimental results are as follows: 
(1) \sysname's attack diagnosis under multi-sensor SDAs achieves a 96\% true positive (TP) rate and a 5\% false positive (FP) rate. 
(2) \sysname effectively masks FPs from the attack detector, leading to a 4X reduction in gratuitous recovery activations in contrast to three other baseline diagnosis approaches. 
(3) \sysname successfully recovers RVs from multi-sensor SDAs in 93\% of cases, resulting in over 2X reduction in the RV's deviations from the set path, and a 2.5X reduction in mission delays compared to a worst-case recovery technique. 
(4) \sysname also mitigates stealthy attacks and achieves $100\%$ mission success under such attacks, and (5) \sysname incurs modest overheads in both runtime and memory when executed on real RVs, and is able to recover them from SDAs without incurring any crashes. 

\section{Background and Threat Model}
\label{sec:background}
In this section, we present an RV's state estimation process. 
Then, we present sensor deception attacks (SDA), and the threat model. 

\subsection{State Estimation and Control in RVs}
\label{sec:bg-control}
An RV’s state estimation and control process operates in a feedback loop. It uses the sensor measurements to determine the current physical state of the vehicle, and derive actuator signals for positioning the vehicle in the next state. 
For example, GPS measures the RV's position, gyroscope measures angular velocity, accelerometer measures velocity and acceleration, magnetometer measures the heading direction, and barometer measures the altitude. 
In addition, RVs use sensor fusion techniques like Extended Kalman Filter (EKF)~\cite{ekf} to enhance the physical state estimations by fusing measurements from sensors with a model of the systems's dynamics. 
Typically, a PID (Proportional Integral Derivative) controller is used for the RV's position, velocity, and orientation control.

\begin{figure}[!ht]
	\includegraphics[width=0.45\textwidth]{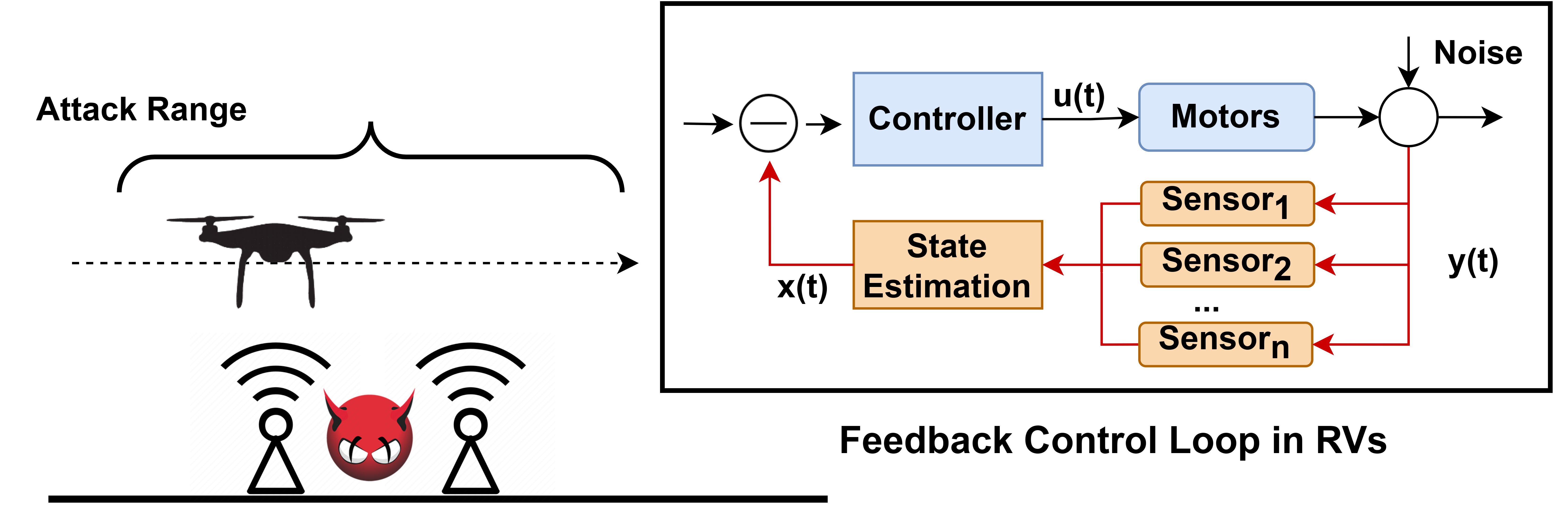}
	\centering
	\caption{State Estimation and Feedback Control in RVs. SDA influencing RV's state estimation and control.}
	\label{fig:control}
\end{figure}


\subsection{Sensor Deception Attacks}
Physical attacks that manipulate sensor measurements from an external source via physical channels have been demonstrated in RVs. 
\eg gyroscope measurements can be manipulated through acoustic noise injection~\cite{gyroscopespoofing, injected-delivered}, and GPS measurements can be manipulated by transmitting false GPS signals~\cite{gpsspoofing1, tractor-beam}. 
Such attacks are launched by injecting false data to sensor~\cite{opticalsensorspoofing, stealthy-attacks, choi, savior}. 

In this paper, we consider a strong form of physical attacks against RVs - we call these {\em Sensor Deception Attacks} (SDA) --  in which, one or more sensors in the RV are compromised simultaneously. 
Any possible combination of sensors can be targeted by an SDA, even up to all the sensors.
An SDA is launched by injecting malicious signals to the RV's sensors. 
For example, an adversary can set up a fake GPS emitter, acoustic signal injectors, and magnetic signal injectors at a geographical location, 
and target the GPS, gyroscope, accelerometer, and magnetometer sensors simultaneously during an RV mission.
Such an SDA can be debilitating for the RVs, as it {\em simultaneously} corrupts the physical properties measured by different types of sensors on the RV (\eg position, velocity, acceleration, angular velocity, heading direction etc.)

SDAs targeting multiple sensor types have been demonstrated against bare metal systems~\cite{sensor-confusion} and self-driving cars~\cite{cam-lidar}. 
For drones, Nashimoto~\etal \cite{sensor-confusion} showed that an adversary can target gyroscope, accelerometer, and magnetometer sensors at once through acoustic and magnetic signal injection to influence state estimation even in the presence of sensor fusion (EKF). 
We later show that such an attack can crash a drone almost immediately. 

Attackers can also exploit knowledge of an  RV's attack detection strategy to launch stealthy attacks~\cite{stealthy-attacks}. 
Stealthy attacks inject controlled sensor manipulations while evading detection,  
and can cause significant disruptions when performed over long durations~\cite{stealthy-attacks-2}.  

\subsection{Threat Model and Assumptions}
\label{sec:threat-model}
We focus on physical attacks that maliciously perturb one or more of the RV's sensors' measurements. 
To launch the attacks, attackers can deploy signal emitters in locations of their choosing. 
The attacker can also manipulate any number of RV sensors during the RV mission. 
They can also inject sensor manipulation in a controlled manner to launch stealthy attacks. 
Attacks targeting RV's software components or communication channels are considered out of scope as they can be handled by existing techniques~\cite{cfi}. 

\smallskip
\noindent
\textbf{Assumptions}. We make three assumptions. First, we do not consider obstacles in the RV's path, which is in line with almost all the prior work in this area~\cite{recovery-lqr, srr-choi, pid-piper}). 
This is because the RV's trajectory planning can be equipped with complementary obstacles or collision avoidance components if obstacles are expected in the mission~\cite{obstacle-avoidance}. 
Second, we assume that the RV's starting position is an attack-free zone. 
This is because the RV needs to gather some attack-free historical state information to enable diagnosis and recovery (details in Section~\ref{sec:implementation}). 
Finally, we assume that the attacker cannot control the RV's environmental conditions, e.g., wind.

\section{Related Work and Motivation}
\label{sec:related-work}
In this section, we first present the related work and highlight its limitations. Then, we present a motivating example for introducing attack diagnosis guided targeted attack recovery from SDAs.

\subsection{Related Work}

\textbf{Attack Detection.}
Many attack detection techniques have been proposed for RVs~\cite{choi, savior, pid-piper, deep-sim, m2mon}, and other cyber-physical systems (CPS)~\cite{ids_mutants, ids_noise_matter, ids_autoencoder_swat} that detect attacks based on either invariants or model estimations.
However, all of these techniques focus only on attack detection which, by itself, is not enough for securing RVs. 
All three components attack detection, diagnosis, and recovery are necessary for effectively mitigating SDAs against RVs.

\smallskip
\noindent
\textbf{Attack Recovery.}
We classify prior work in attack recovery on RVs into two categories as follows: (1) \mbr recovery, and (2) \cbr recovery techniques.

\mbr recovery techniques rely on a model of the RV to derive corrected sensor or actuator signals to recover RVs from attacks. 
For example, Software Sensor-based Recovery (SSR)~\cite{srr-choi} emulates the physical sensor signals using virtual sensors, and derives recovery control actions.  
Another example is PID-Piper~\cite{pid-piper}, which uses a machine learning (ML) based feed-forward controller (FFC) to derive robust actuator signals under attacks. 

On the other hand, \cbr recovery techniques isolate sensors upon attack detection, and instead use the historic trustworthy states of the system to determine robust control actions and recover RVs. 
Kong \etal~\cite{pcb-kong} use the historic states to roll forward the RV to expected states under attack. 
Zhang \etal use a linear programming-based recovery controller~\cite{recovery-lp}, and generate recovery control actions using historic states for bringing a system under attack back to a set target. 
In subsequent work, they 
proposed a linear quadratic regulator (LQR) based recovery controller~\cite{recovery-lqr} which improves the recovery control actions and ensures that the system remains in a safe state after recovery. 


\smallskip
\noindent
\textbf{Limitations in Existing Work and Research Gap.}
A fundamental issue with Model-based approaches ~\cite{srr-choi, pid-piper} is that they are based on a predictive model of the RV, and hence they tolerate an error margin (approximation error) between model estimations and the RV's real behavior. 
The error margin is small enough under single sensor attacks to derive robust control actions for recovering the RV. 
However, when multiple sensors are attacked simultaneously, the model's approximation error increases. 
Thus, both SSR and PID-Piper are ineffective under multi-sensor attacks.

In contrast, \cbr techniques derive recovery control actions aimed at restoring the RV to a predefined state, instead of merely tolerating attack-induced manipulations.
Thus, they prove superior in safeguarding RVs from attacks.   
However, \cbr recovery methods assume all the sensors are targeted simultaneously ~\cite{pcb-kong, recovery-lp, recovery-lqr} and perform a worst-case recovery irrespective of the actual number of sensors targeted. 
We show later that the worst-case assumptions lead to overly conservative or aggressive maneuvers of the RV which result in disruptions and a failed recovery, when only a subset of sensors are attacked. 

\smallskip
\noindent
\textbf{Scope of this Paper.}
Our goal is to bridge the gap between \mbr and \cbr techniques by enabling attack recovery methods to handle single or multi-sensor SDAs, even those that target a subset of sensors.  
To achieve this, we propose an attack diagnosis technique that integrates seamlessly with existing attack detection~\cite{pid-piper, savior} and recovery techniques~\cite{recovery-lqr}. 
Upon attack detection, our proposed attack diagnosis identifies the targeted sensors and isolates them from the RV's feedback control loop. 
Subsequently, recovery control actions are generated corresponding to the compromised sensors, eliminating the worst-case recovery assumptions when only a subset of sensors is targeted.

Further, to address attack detection delays caused by stealthy attacks which can remain undetected for a prolonged duration and impact recovery process, we design a method to determine the maximum duration a stealthy attack can remain undetected. 
We then synchronize attack detection, diagnosis, and recovery techniques in a unified system to prevent disruptions due to stealthy attacks. 


\subsection{Motivation}
\label{sec:motivation}
We performed an experiment and launched SDAs targeting a subset of the RVs sensors to highlight the gaps in the current \cbr attack recovery techniques. The details of the experimental setup are explained later (Section~\ref{sec:experiments}). 
We used the state-of-the-art \cbr recovery technique, which employs a Linear Quadratic Regulator (LQR)~\cite{recovery-lqr} for deriving recovery control actions \ie corrective maneuvers for safeguarding the RV.  
Henceforth, we refer to this method as \lqr. 
Recall that when recovery is activated \lqr isolates all the sensors from the RV's feedback control loop, and assumes a worst case recovery scenario. 

\begin{figure}[!ht]
	\centering
	\includegraphics[width=0.40\textwidth]{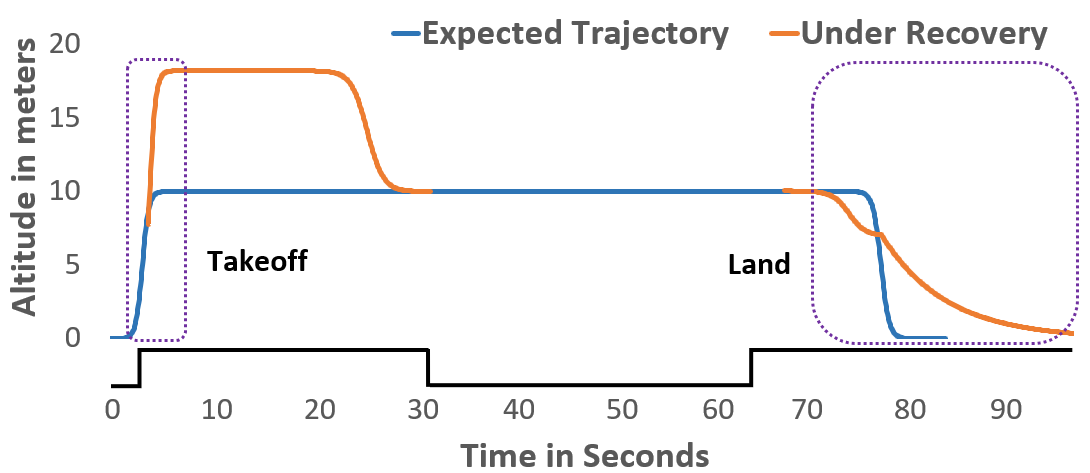}
	\caption{\lqr's recovery from SDA targeting 2 sensors simultaneously. 
    The bottom graph shows the SDA duration. Two instances of SDAs launched from t=5-30s and t=65-95s. 
    }
	\label{fig:motivation}
\end{figure}

In this experiment, we used a Pixhawk drone (details in Section~\ref{sec:experiments}), and we launched SDAs targeting the RV's GPS and accelerometer sensors. 
We deliberately target two out of the five sensors of the drone to study \lqr's recovery when only a subset of sensors is targeted by SDAs.  
We assume the drone has an attack detector~\cite{pid-piper} that raises an alert upon an SDA, activating attack recovery. 
Figure~\ref{fig:motivation} shows the drone with \lqr's recovery under the SDAs. 

The first instance of SDA was launched between 5 to 30s, during the takeoff phase of the RV mission. 
As \lqr assumes a worst-case recovery it disengages the inputs from all the sensors. 
Then, using the historic states it derives recovery control actions to set the RV's altitude at $10$m as per the mission trajectory. 
However, due to the absence of real-time sensor feedback, \lqr erroneously believes the RV remains below the 10-meter target altitude. 
Hence, \lqr triggers an aggressive maneuver, propelling the RV to an elevation of 18 meters, a substantial deviation from its expected trajectory. 
While \lqr eventually steers the drone back on course, it incurs significant deviations and prolongs the mission by 15s. 

The second instance of SDA was launched from t=65 to 95s, during the landing phase of the RV mission.
\lqr again performs a worst-case recovery, this time exhibiting an overly conservative approach in deriving recovery control actions. 
This is because \lqr lacks information regarding the rate at which the RV is descending due to the absence of real-time sensor feedback.
Consequently, \lqr extended the RV's landing duration by approximately 20 seconds. 
Furthermore, it resulted in an imprecise landing with the RV touching down 15m away from the set destination, which is a deviation much larger than the standard GPS offset of $\approx$ 5m. 

\smallskip
\noindent
\textbf{Why attack diagnosis guided targeted attack recovery?} 
The above experiment highlights the limitations of the state-of-the-art \cbr recovery technique that undertakes a worst-case recovery strategy, irrespective of the actual number of sensors targeted by an SDA.
This results in overly conservative or aggressive maneuvers, which is particularly undesirable during critical phases like drone takeoff or landing, and can lead to mission disruptions or failure.
Recall that \cbr techniques operate under the assumption that all the sensors are under attack regardless of the actual number of sensors targeted.  
Due to the worst case assumptions, \cbr recovery techniques fail to leverage feedback from the uncompromised sensors, thereby limiting their ability to improve recovery control actions.
While worst-case recovery outcomes may be unavoidable if an SDA targets all the sensors, targeted attack recovery can prevent unnecessary recovery failures when the attack targets only a single sensor or a subset of sensors. 

{\em Our goal is to enhance the effectiveness of \cbr techniques by introducing targeted attack recovery strategies guided by attack diagnosis.}
Our approach eliminates the need for worst-case assumptions, and enables deriving recovery actions corresponding to the sensors targeted by an SDA.

\noindent
\textbf{Why a unified framework?}
\cbr techniques heavily rely on the RV's historical behavior to estimate safe recovery control actions. 
However, the persistence of attack detection delay poses a significant challenge as it may lead to erroneous recovery actions. 
This can be addressed by integrating attack detection and recovery techniques, and ensuring that attack-free historical information is used for deriving recovery actions. 
Furthermore, multi-sensor attacks introduce significant challenges for existing \cbr recovery techniques. 
Without attack diagnosis, existing recovery techniques are confined to worst-case assumptions. 
The integration of attack diagnosis with attack recovery enables targeted recovery actions, improving recovery success rates. 

\section{Design}
\label{sec:design}
\begin{figure}[!ht]
	\includegraphics[width=0.475\textwidth]{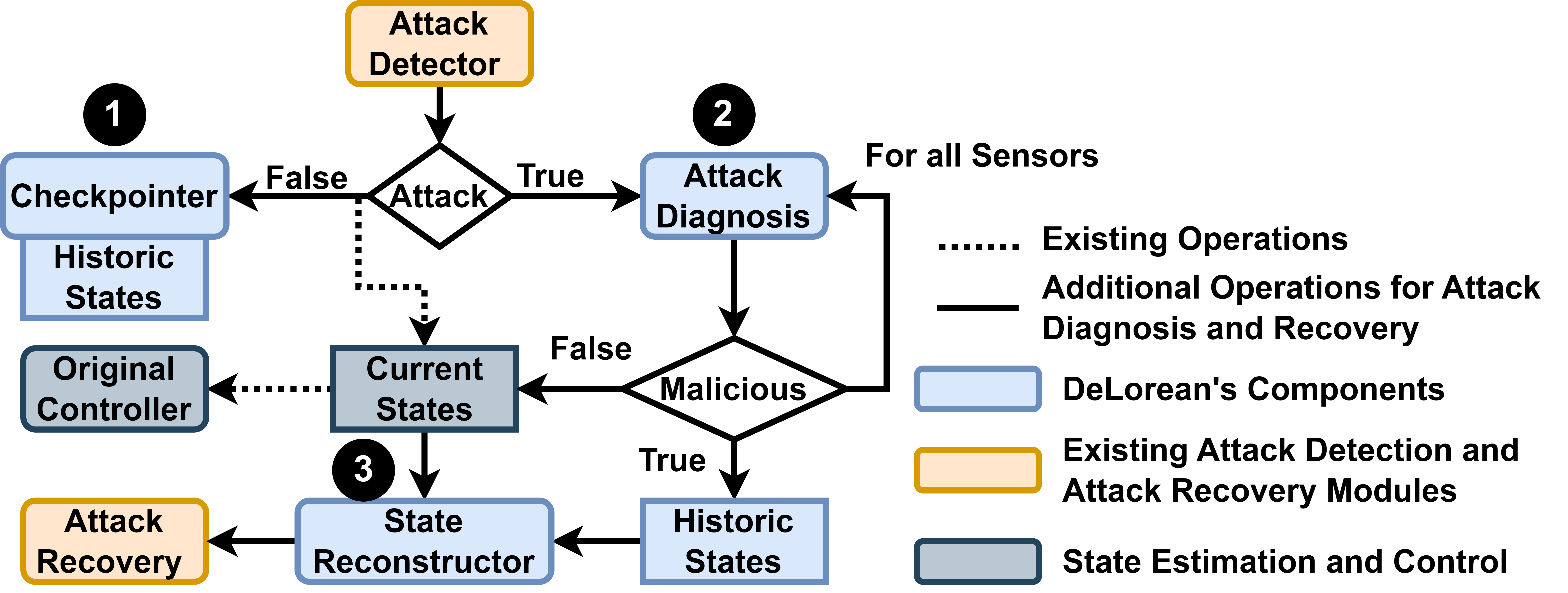}
	\centering
	\caption{Sequence of operations in attack detection, diagnosis, and recovery. These operations run onboard the RV. }
	\label{fig:delorean-flow}
\end{figure}

Figure~\ref{fig:delorean-flow} shows the \sysname framework running onboard the RV. 
\sysname uses an existing attack detector to detect attacks, and performs diagnosis to identify sensors targeted by the attack. 
There are many existing attack detection techniques for RVs~\cite{choi, savior, pid-piper}. 

\begin{figure}[!ht]
	\includegraphics[width=0.475\textwidth]{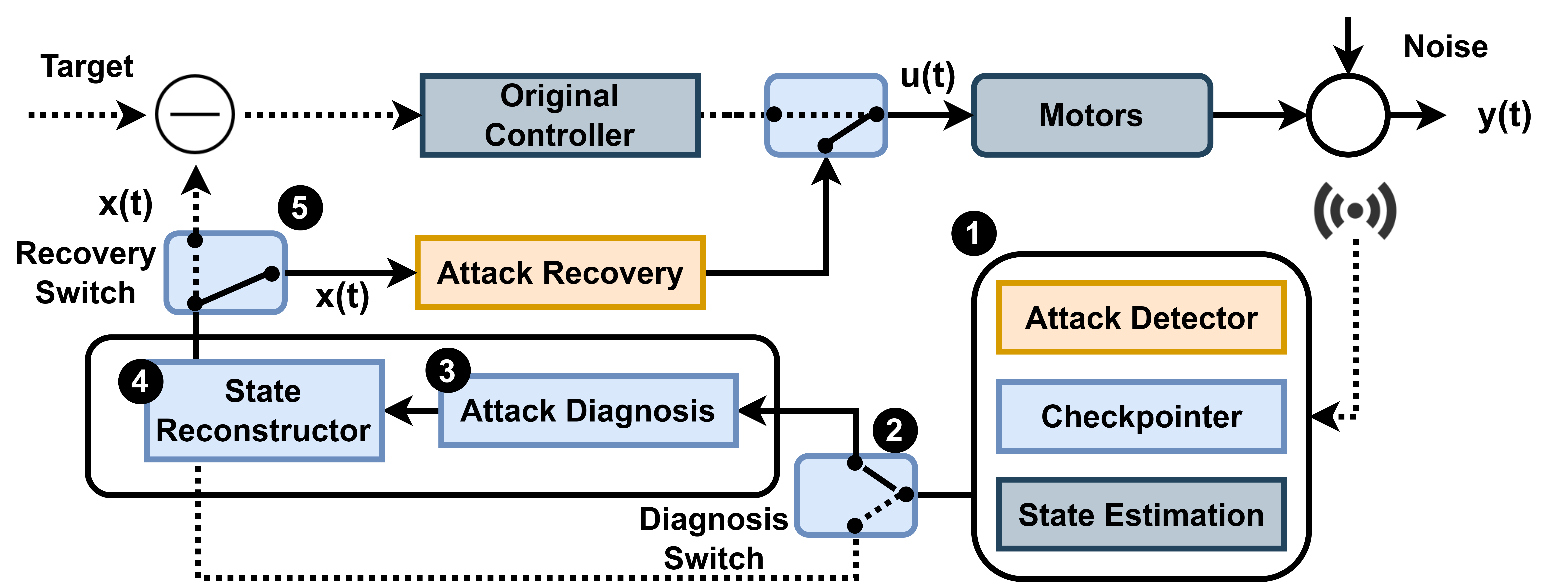}
	\centering
	\caption{Feedback control loop with recovery modules}
	\label{fig:hsr-design}
\end{figure}

\sysname activates diagnosis when the attack detector indicates an attack.
After identifying targeted sensors through diagnosis, \sysname activates attack recovery to apply targeted recovery only for sensors that are malicious. 
Our framework is designed to be versatile and can integrate any \cbr recovery technique~\cite{recovery-lp, recovery-lqr, pcb-kong}. 
Note however that, none of the prior \cbr recovery techniques offer targeted attack recovery. 
We design a checkpointing strategy to record attack-free historic states and a state reconstruction approach that accurately represents the RV's current states under SDAs. 
These supporting components are designed to incorporate the outcomes of attack diagnosis, and enable \cbr techniques to perform targeted recovery actions.

\sysname's operations are shown in Figure~\ref{fig:delorean-flow}.
\sysname consists of three main operations as follows.
(i) \textbf{Historic States Checkpointing}: \sysname records trustworthy historic states from an attack-free phase of the mission in a sliding window. These historic states are used to derive recovery actions. 
(ii) \textbf{Attack Diagnosis}: Under an attack, \sysname identifies the targeted sensors 
to selectively derive recovery control actions.
(iii) \textbf{State Reconstruction}: \sysname reconstructs the RV's states from trustworthy historical states and uncompromised sensor measurements (available when SDAs target a subset of sensors) to obtain an accurate representation of the RV's current states despite the attack induced manipulations.
These reconstructed states are used by the attack recovery technique to derive the recovery control actions. 

Figure~\ref{fig:hsr-design} shows the components of \sysname in RV's feedback control loop. 
As shown in Figure~\ref{fig:delorean-flow}, \sysname's first operation, \emph{Historic State Checkpointing} is executed in the absence of attacks. 
\circlenum{1} A canonical attack detector 
uses the sensor measurements $y(t)$ to derive the RVs next physical state $x'(t)$, and compares it with the values derived by the state estimation module $x(t)$.
If the residual  $r = |x'(t) - x(t)|$ exceeds a predefined threshold, that indicates an attack~\cite{pid-piper, choi, savior}.
Once an attack is detected, \circlenum{2} the {\em Diagnosis Switch} activates \sysname's second operation, \circlenum{3} {\em Attack Diagnosis}. 
This step identifies the sensors targeted by the attack, and isolates the targeted sensors from the feedback control loop, preventing the attack induced sensor manipulations from propagating and causing cascading errors in the RV's feedback control loop. 
Based on attack diagnosis, \sysname's third operation \circlenum{4} \emph{State Reconstructor} utilizes measurements from uncompromised sensors and historical states for compromised sensors to estimate accurate state representation of the RV.
Finally, \circlenum{5} the {\em Recovery Switch} forwards the reconstructed state estimates to the attack recovery component for deriving recovery control actions. 

\subsection{Attack Diagnosis}
\label{sec:attack-diagnosis}
The first step in \sysname is to identify the targeted sensors once an attack is detected. 
Attack induced sensor manipulations corrupt the RV's physical states that are estimated using the targeted sensors. 
For example, when the gyroscope sensor is under attack, it leads to the corruption of the RV's Euler angles (roll, pitch, yaw), and angular velocity. Table~\ref{tab:states2sensor} shows the states to sensor mapping.

\begin{table}[!ht]
	\centering
	\footnotesize
	\caption{Physical states and corresponding sensors in RVs}
	\begin{tabular}{l|l|l|l|l|l}
		\hline
		\textbf{Sensors} & \textbf{GPS}                                                                              & \textbf{Gyroscope}                                                                                                      & \textbf{Accel}                     & \textbf{Barometer} & \textbf{Magnetometer}      \\ \hline
		\textbf{States}  & \begin{tabular}[c]{@{}l@{}}$x$, $y$, $z$, \\ $\dot{x}$, $\dot{y}$, $\dot{z}$\end{tabular} & \begin{tabular}[c]{@{}l@{}}$\phi$, $\theta$, $\psi$, \\ $\omega_\phi$, $\omega_\theta$, $\omega_\psi$\end{tabular} & $\ddot{x}$, $\ddot{y}$, $\ddot{z}$ & $z$           & $x_m$, $y_m$, $z_m$ \\ \hline
	\end{tabular}
	\label{tab:states2sensor}
\end{table}

We observe that in attack-free segments of the mission, the error between the RV's past and present states is relatively low \ie, 
$e_{i_t} = |state_t - state_{t-1}|$ $\ll \delta$ and $e_{i_{t-1}} = |state_{t-2} - state_{t-3}| \ll \delta$, where $i$ $\in$ PS (all the physical states of the RV, shown in Equation~\ref{eqn:all-states})
However, under attack, the error $e_i$ increases \ie, $e_i \gg \delta$~\cite{choi,savior,pid-piper}.
Therefore, our approach for attack diagnosis is to monitor the inflation in error $e_i$ in four consecutive time steps and find the probability that the corresponding sensor is under attack given the observed errors (including both steady and non-steady state errors)
\begin{equation}
	\thinspace
	PS = x, y, z,\dot{x}, \dot{y}, \dot{z}, \ddot{x}, \ddot{y}, \ddot{z}, \phi, \theta, \psi, \omega_\phi, \omega_\theta, \omega_\psi, x_m, y_m, z_m
	\label{eqn:all-states}
\end{equation}

However, error inflation in the RV's physical states can also occur due to the RV's physical dynamics in the mission (\eg mode changes), and not only due to attacks. 
Thus, it is important to find the root cause of the error inflation in RV's physical states. 
We use factor graphs (FG) to monitor the error inflation in physical state.  
FGs are probabilistic graphical models that allow expressing conditional relationships between variables~\cite{factor-graph}.
In particular, we use FG to represent the causal relationship between the observed error $e_i$ and find the probability that the corresponding sensor is under attack \ie the outcome $s_i$. 
We consider binary outcomes for the sensors \ie {\em malicious} (targeted by SDA) or {\em benign} (attack-free). 

The causal relationship between $e_i$ and $s_i$ can be expressed as a conditional probability problem \ie $P(s_i|e_i)$. 
There are two ways to calculate conditional probability: 
(1) Using a joint probability distribution over the observed $e_i$, and the probable outcomes $s_i$ (\ie benign or malicious sensors). 
However, expressing the joint probability distribution at runtime for binary outcomes will have high computation and storage overheads. 
Further, as the mission progresses, the time taken to calculate probabilities will increase exponentially. 
Therefore, this approach is not practical. 

(2) Using a Bayesian approach, we can calculate $P(s_i|e_i)$ based on the values of known probabilities of $P(e_i|s_i=malicious)$ and $P(e_i|s_i=benign)$. 
To calculate $P(s_i|e_i)$, we will need to run a large number of experiments and profile the observed error $e_i$ in the RV's physical states under attacks as well as in the absence of attacks to calculate $P(e_i|s_i=malicious)$ and $P(e_i|s_i=benign)$. 
This will also not work in our case due to two reasons:  First, it is difficult to know how many observations are enough.  
Second, if we observe an error during inference that was not observed during profiling, the Bayesian approach will assign it a zero probability (\ie, Zero Frequency problem).
Thus, even if a sensor is under attack, the Bayesian approach will wrongly conclude the sensor is benign.

In contrast to the above approaches, FGs allow fine-grained representation of the complex joint probability distribution between variables through a product of smaller probability distributions. 
Thus, they incur low computation and storage overheads. Further, instead of relying on profiling, FGs use functions (\ie, factor functions) to express conditional relationships. 
Thus, FGs can accommodate unseen observations ($e_i$), and do not suffer from the Zero Frequency problem. 
This is why we use FGs for attack diagnosis. 

\begin{figure}[!ht]
	\centering
	\includegraphics[width=0.40\textwidth]{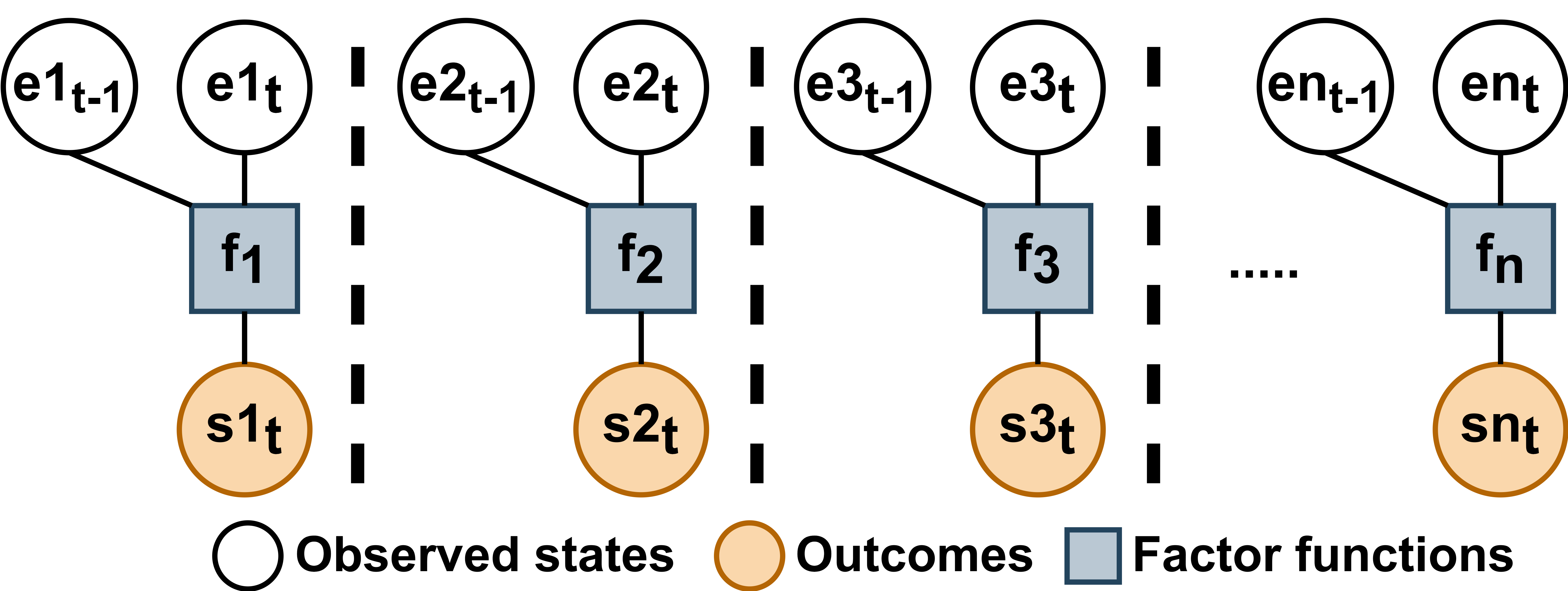}
	\caption{Factor graph for diagnosing malicious sensors under an SDA given the observed errors in RV's physical states.}
	\label{fig:factor-graphs}
\end{figure}

Figure~\ref{fig:factor-graphs} shows an example of FG-based diagnosis.  
The FG represents the relationships between error $E = (e_1,.., e_n)$ and the probable outcome $S = (s_1,.., s_n)$ using a bipartite graph, where $e_i$ and $s_i$ are observed error and outcomes of the RV's physical states (Equation~\ref{eqn:all-states}).
Factor functions $F = (f_1,.., f_n)$ define relationships between observed errors ($e_{i_{t-1}}, e_{i_t}$) and probable outcomes $s_i$.
Using FGs, we calculate the probability of each physical state $s_i$ being malicious given the observed error $e_i$ \ie $P(s_i = malicious|e_i)$. 
If certain physical states are found to be malicious, we deem the corresponding sensors (Table~\ref{tab:states2sensor}) to be under attack.  

\noindent \textbf{Factor Graph Construction.}
The steps involved in constructing the FGs and using FGs for attack diagnosis are as follows. 

{\em Step 1 - Express the causal relationships between errors in RV's physical states and the probable outcome using factor functions.}
To express such a relationship, the first step is to determine the safe and unsafe ranges of the error $e$. 
We run attack-free missions and collect mission traces, and we observe the variation in error $e$.  
Recall that in an attack free phase, the error is low: $e+\epsilon$, where $\epsilon$ is environmental noise. 
We deem $e$ to be in the safe range if $0 < e < \delta$, where $\delta = median(e) + k * stdev(e)$. 
The value of $\delta$ is RV-specific, and it is derived empirically  
(details in Section~\ref{sec:results}). 

The factor functions rely on the $\delta$ values of the respective physical states to output a discrete value representing the possible outcome for each sensor (Equations~\ref{eqn:type-1}). 
In particular, the factor function expresses the relations if $e_{i_{t-1}} > \delta$ and $e_{i_t} > \delta$, then what is the most probable outcome for the sensor (malicious or benign)?
Note that we monitor errors in the past four states ($e_{i_t} = |state_t - state_{t-1}|$, $e_{i_{t-1}} = |state_{t-2} - state_{t-3}|$) to disregard transient error inflation and benign changes in physical dynamics in the mission. 

\noindent
\begin{equation}
	\footnotesize
	f_i(e_{i_{t-1}}, e_{i_t}, s_{i_t}) = \left\{\begin{matrix}
		1 ,& e_{i_t} > \delta \text{~and~} e_{i_{t-1}} >\delta \text{~and~} \text{if~} s_t = malicious  \\ 
		0 ,&  otherwise  
	\end{matrix}\right.
	\label{eqn:type-1}
\end{equation}

{\em Step 2 - Construct per sensor factor graphs}.
We repeat Step-1 and construct FGs expressing relationships between the observed errors $e_i$ and the probable outcomes $s_i$ for all the physical states 
corresponding to all the sensors of the RV.
The factor functions shown in Equation~\ref{eqn:type-1} is used for this purpose (only the value of $\delta$ changes, which is empirically determined). 

{\em Step 3 - Perform inference on the factor graphs to diagnose the targeted sensors.}
We feed the physical states derived from sensor measurements to the per-sensor FGs to determine the targeted sensors.  
We calculate the probability of $s_{i_t}$  = malicious, for all the physical states PS (Equation~\ref{eqn:all-states}) given the observed $ e_{i_{t-1}}, e_{i_t}$. 
The joint probability distribution can be factorized as:
\begin{equation}
	\thinspace
	P(E_t, S_t) = \prod_{i=1}^{n} f_i(e_{i_t}, e_{i_{t-1}}s_{i_t}), \, {i \in PS}
\end{equation}
Thus, at each time $t$, the conditional probability is 
\begin{equation}
	\thinspace
	P(s_{i_t} = \text{malicious} |e_{i_t}) = \prod_{i=1}^{n} f_i(e_{i_t}, e_{i_{t-1}}s_{i_t}), \, {i \in PS}
\end{equation}
We assume that each sensor is initially equally likely to be in a benign or malicious state, so $P(s_i = benign)$ = 0.5 and $P(s_i = malicious)$ = 0.5.
We use Maximum Likelihood Estimation (MLE) to find the likelihood of each possible outcome of $s_{i}$ given the observed errors $e_i$.
Because the observed $e_i$ inflates under attacks, the $P(s_i = malicious) > P(s_i=benign)$. 
We deem those states to be malicious where $P(s_{i_t} = malicious)> 0.5$, and deem the corresponding sensors (Table~\ref{tab:states2sensor}) as being under attack.  

\subsection{Historic States Checkpointing}
\label{sec:hsr}
\begin{figure}[!ht]
	\centering
	\begin{subfigure}{.37\linewidth}
		\centering
		\includegraphics[width=\linewidth]{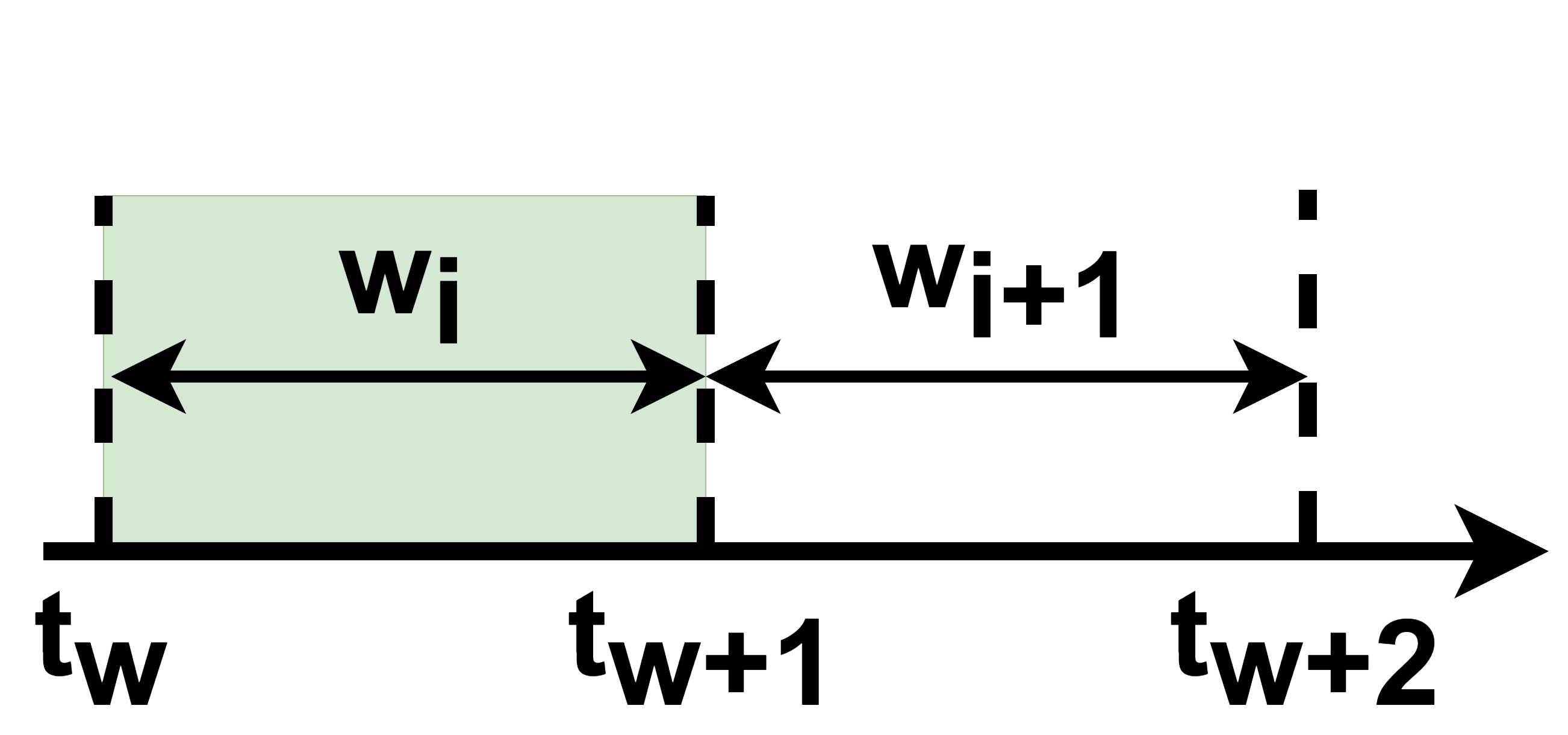}
		\caption{}
		\label{fig:window-no-attack}
	\end{subfigure}%
	\hspace{1.5 em }
	\begin{subfigure}{.47\linewidth}
		\centering
		\includegraphics[width=\linewidth]{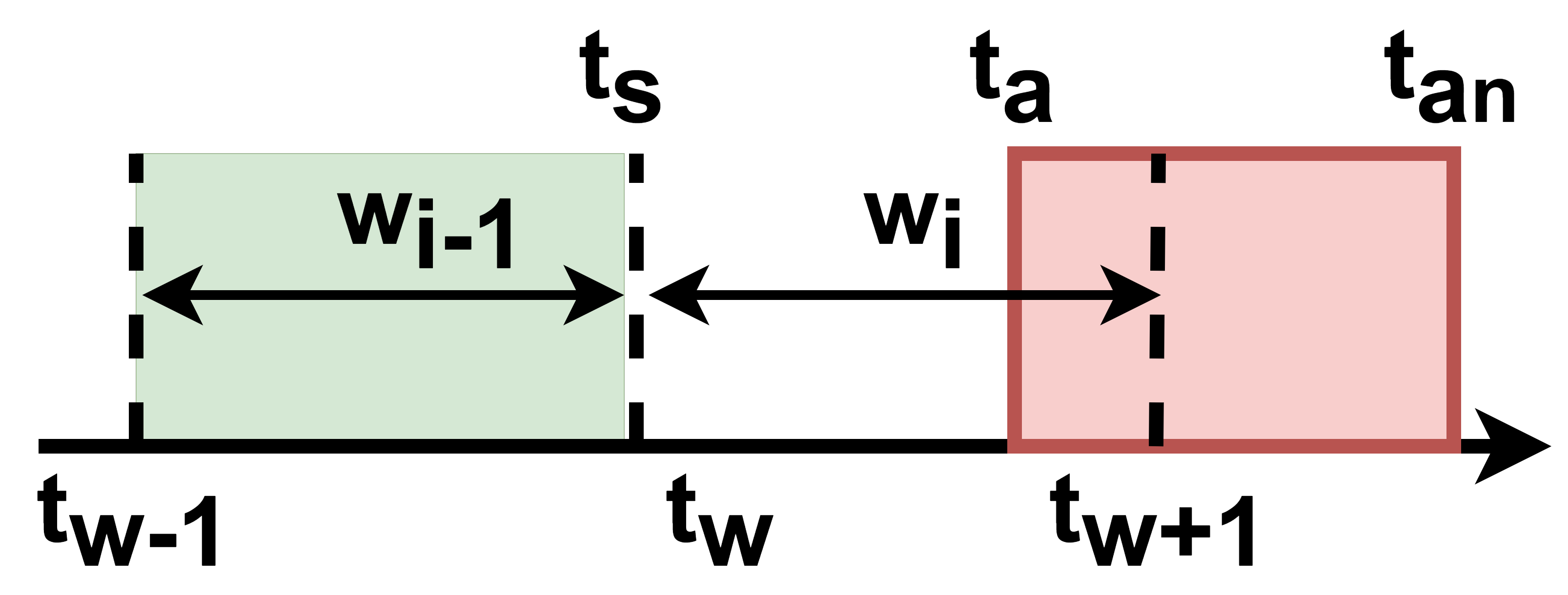}
		\caption{}
		\label{fig:window-attack}
	\end{subfigure}
	\caption{(a) Recording historic physical states in a sliding window. (b) Ensuring historic states are attack-free.}
\end{figure}

When the RV is in an attack free phase, we record the sequence of physical states estimated from all the on-board sensors in a sliding window. 
Figure~\ref{fig:window-no-attack} shows the details. 
The physical states at time $t$ are represented as $x(t)$, and the historic physical states recorded in a window are represented as $HS$ (states shown in Equation~\ref{eqn:all-states}). 

From the start of the RV's mission, we record $HS$ in a sliding window of length $w$.  
If no alert is raised by the attack detector, we save the $HS$ recorded in window $w_i$ and proceed recording states in the next window $w_{i+1}$ as shown in Figure~\ref{fig:window-no-attack}. 
At the end of window $w_{i+1}$, we discard the $HS$ recorded in the previous window $w_i$. 
However, if we detect an attack as shown in Figure~\ref{fig:window-attack}, we stop the recording, and discard the states recorded in the current window $w_{i}$ as these may be corrupted by the attack. 
We use the $HS$ recorded in the previous window $w_{i-1}$ for state reconstruction and recovery. 
These $HS$ are attack-free as no alert was raised by the attack detector in window $w_{i-1}$.

Another challenge is that sensors in RVs are sampled at different rates \eg the sampling rate for gyroscope is 400 Hz, whereas, the sampling rate for Barometer is 100 Hz. 
It is important to align the data streams when recording the historic states. 
To address this problem, we select a single target frequency for recording the $HS$, which is the highest sampling rate of all the sensors. 
We then align the low frequency streams with the high frequency streams by inserting additional data points in the low frequency stream to match the sample points in the high frequency streams. 
In particular, we duplicate the last data point in the low frequency streams based on the ranges of the sample points of the high frequency streams.

\smallskip
\noindent
\textbf{Recovery under Stealthy Attacks.}
However, it is possible for an attack to be stealthy and span multiple sliding windows~\cite{stealthy-attacks}, in which case, it will be challenging to ensure an attack-free $HS$ for recovery.
Such stealthy attacks can inject seemingly benign and controlled bias in the sensor measurements over a long time, thereby evading detection, and causing disruption in the RV's mission. 
Many attack detection techniques such as Savior~\cite{savior} and PID-Piper~\cite{pid-piper} detect stealthy attacks using cumulative sum statistics (CUSUM). 
However, there is necessarily a delay in attack detection \ie the time interval between the onset of the stealthy attack and the time when the attack detector raises an alert. 
Due to the detection delay, \sysname might record corrupted $HS$ for recovery, which it incorrectly deemed safe as no alarms were raised by the attack detector. 
Therefore, to handle stealthy attacks, we empirically choose the window length to be large enough so that stealthy attacks can be detected within a single sliding window.  

\subsection{State Reconstruction}
\label{sec:state-reconstruction}
After identifying targeted sensors through attack diagnosis, we reconstruct the RV's state vector represented as $X'(t)$. 
The goal of state reconstruction is to ensure accurate current state representation 
for the sensors affected by SDAs (referred to as $x_r(t)$), while preserving the state derived from sensors that remained unaffected by the attack at the time when recovery is activated.
Furthermore, the state reconstruction process plays a key role in enabling targeted attack recovery as it facilitates real-time sensor feedback (if available \ie when SDAs target only a subset of the sensors). 

We use the Extended Kalman Filter (EKF)~\cite{ekf} to estimate $x_r(t)$ using the latest trustworthy historic states under SDAs. 
The EKF estimator is constructed offline using the RV's non-linear system dynamics (Appendix~\ref{app:ekf}). 
The parameters of the non-linear equations are learned using a dataset of control actions and sensor measurements. 
We collect the data using our subject RVs (Section~\ref{sec:experiments}). 
We learn an approximate model of the RV $f(x_t, u_t)$ through system identification~\cite{matlab-si}. 
The model parameters are optimized using the least squared error approach, \ie minimize squared error between the model's estimations and the observed values.

Taking the scenarios in Figure~\ref{fig:window-attack} as an example, the latest trustworthy states $x_{t_s}$ are recorded at time $t_s$. 
The RV's states evolve over time as per its system's dynamics~\cite{ekf}. 
For example, at time $t_{s+1}$ the RV's states are calculated as $x_r(t_{s+1}) = f(x_{t_s}, u_{t_s}) + v_{t_s}$, where $u_{t_s}$ and $v_{t_s}$ are control inputs and process noise  respectively. 
The above step is iteratively performed for each time step between $t_s$ and $t_a$ (\ie $t_{s+1}, t_{s+2},.., t_a$) to estimate RV's states $x_r(t_a)$.
The EKF relies on system's dynamics and trustworthy historic states instead of compromised sensor measurements. 

We reconstruct the RV's state vector at $t_a$ when recovery is activated as follows. 
For sensors that remain uncompromised under SDA, we use the states derived from the corresponding sensors, these states are represented as $x_c(t_a) = {x_{c_1},.., x_{c_n}}$. 
We use the trustworthy historic states to estimate states corresponding to the sensors that are targeted by the SDA, these states are represented as $x_r(t_a) = {x_{r_1},.., x_{r_n}} $.
We reconstruct RV's state vector by selectively combining the states derived from uncompromised sensors and estimated states for the compromised sensors as $X'(t_a) = [x_c(t_a), x_r(t_a)]$ to obtain accurate current state representations.
$X'(t_a)$ is the initial system state of recovery.

\section{Experimental Setup and Parameters}
\label{sec:experiments}
In this section, we present our experimental setup, metrics, and attack parameters for evaluating \sysname.  We then present \sysname's parameters derived from our experiments~\footnote{We have made our code and data publicly available at \url{https://github.com/DependableSystemsLab/DeLorean}}. 

\begin{figure}[!ht]
\centering
\begin{subfigure}{.24\linewidth}
	\centering
	\includegraphics[width=0.80\linewidth]{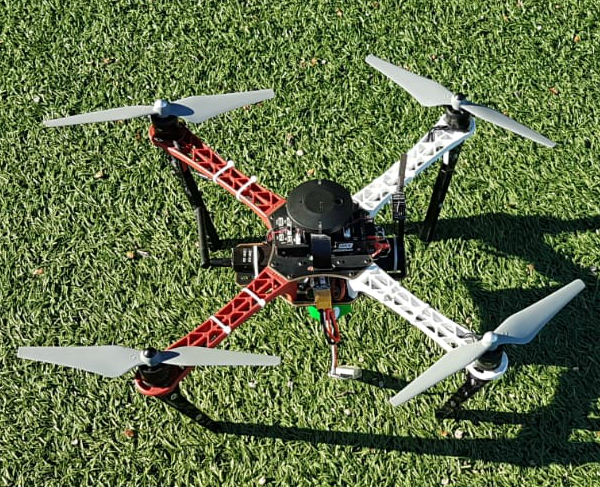}
\end{subfigure}%
\begin{subfigure}{.24\linewidth}
	\centering
	\includegraphics[width=0.90 \linewidth]{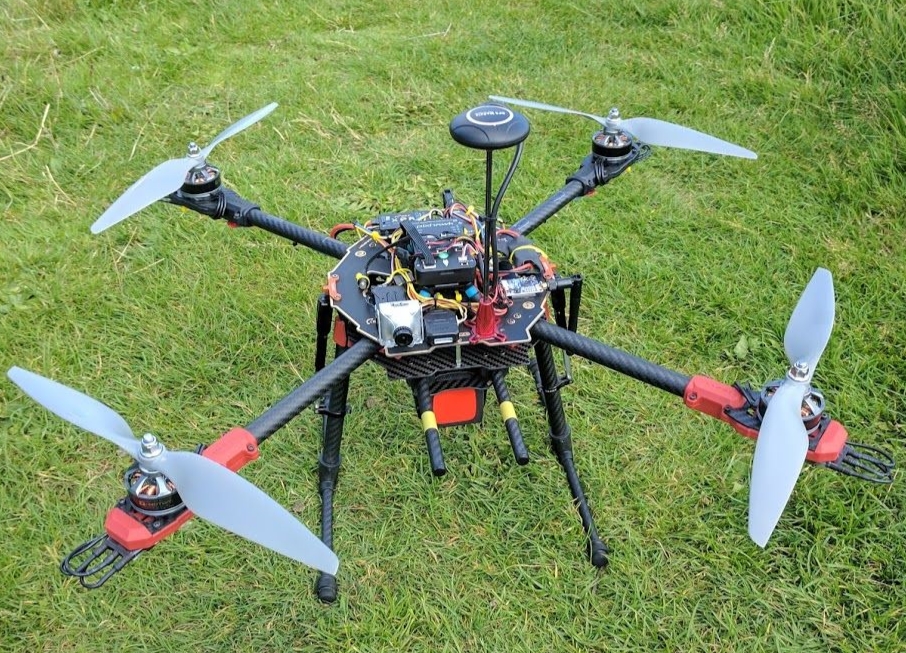}
\end{subfigure}
\begin{subfigure}{.24\linewidth}
	\centering
	\includegraphics[width=0.85\linewidth]{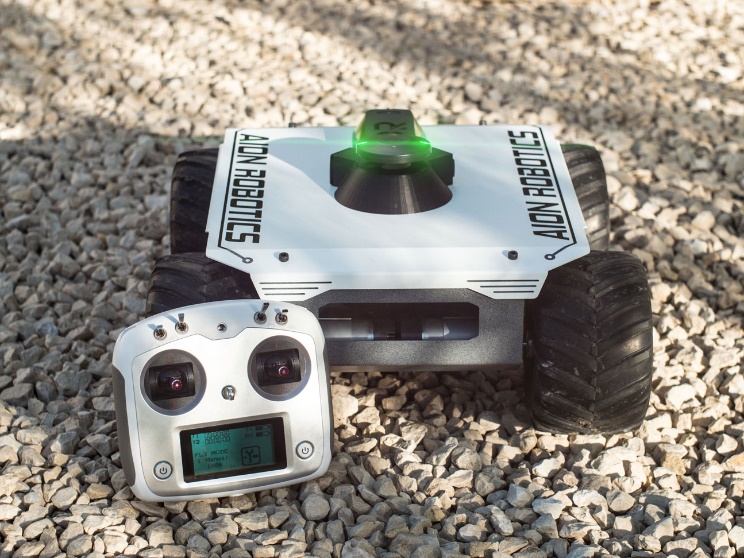}
\end{subfigure}
\begin{subfigure}{.24\linewidth}
	\centering
	\includegraphics[width=0.75 \linewidth]{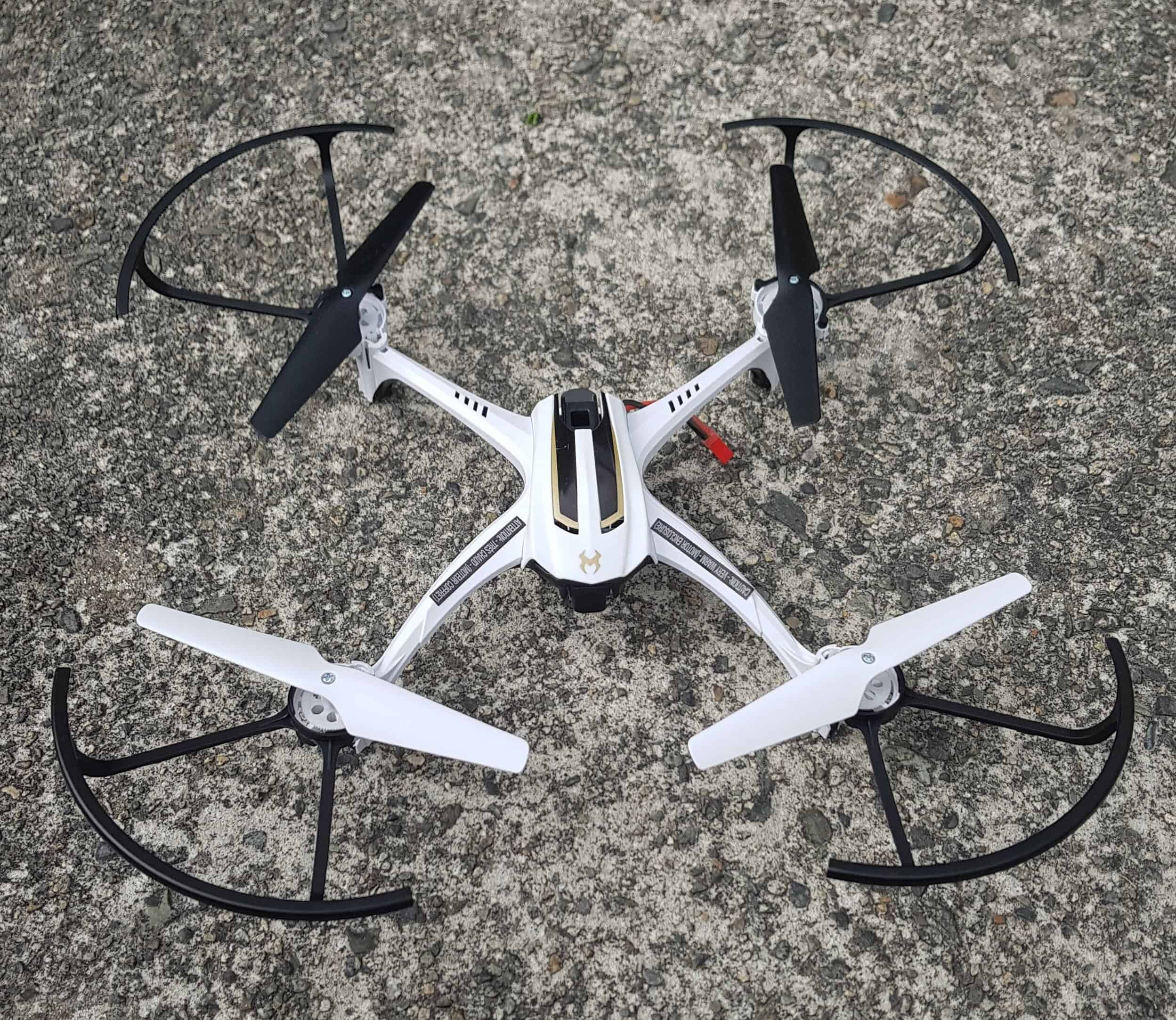}
\end{subfigure}
\caption{Real RV Systems (a) Pixhawk drone, (b) Tarot drone, (c) Aion Rover, (d) Sky Viper drone.}
\label{fig:real-rv}
\end{figure}

\smallskip
\noindent
\textbf{Subject RVs:}
To evaluate \sysname, we use six RV systems. Four of these are real RVs. These are shown in Figure~\ref{fig:real-rv} 
(from the left)  (1) Pixhawk based DIY drone~\cite{Pixhawk} (Pixhawk drone), (2) Tarot 650 drone~\cite{tarrot} (Tarot drone)
(3) Aion R1 ground rover~\cite{aion} (Aion rover), and (4) Sky Viper Journey drone~\cite{skyviper} (Sky-viper drone). 
The first three RVs are based on the Pixhawk platform~\cite{Pixhawk}. 
The Sky-viper drone is based on an STM32 processor. 
These RVs are all equipped with at least one of the following five sensors: GPS, gyroscope, accelerometer, barometer and magnetometer, but 
each RV has different numbers of individual sensors (Table~\ref{tab:attack-recovery-implementation}).  
The other two RV systems are simulated RVs, (4) Ardupilot's quadcopter (ArduCopter), and (5) Ardupilot's ground rover~\cite{ardupilot} (ArduRover). 
We use the APM SITL~\cite{ardupilot}, and Gazebo~\cite{gazebo} platforms for  simulations. 

\noindent
\textbf{RV Missions:}
We run a diverse set of 340 missions with varying mission duration, and mission distances. 
A mission is an autonomous navigation path of the RV, including a starting point, intermediate waypoints, and a designated endpoint.
We simulate a wind flow between 0-10 m/s to vary the environmental conditions.
The mission emulates various real-world RV missions: (1) a last mile delivery drone~\cite{upsdrugdelivery} (straight line path, polygonal path), (2) drones used for surveillance or agriculture~\cite{dji-agriculture} (circular or polygonal path),  and (3) rovers deployed in warehouse management~\cite{kiva} (polygonal paths).
Appendix~\ref{app:mission-types} discusses the RV mission paths in depth.


\begin{table}[!ht]
	\footnotesize
	\caption{Subject RVs in Evaluation, Number of Sensors, Attack parameters. P: Pixhawk drone, T: Tarot drone, 
		S: Sky-viper drone, AC: ArduCopter, R1: R1 rover, AR: ArduRover}
	\begin{tabular}{l|llllll|ll}
		\hline
		\multirow{2}{*}{\textbf{Sensor Type}} & \multicolumn{6}{l|}{\textbf{Number of Sensors}}                                                                                                                                                        & \multicolumn{1}{l|}{\multirow{2}{*}{\textbf{\begin{tabular}[c]{@{}l@{}}Bias\\ Values\end{tabular}}}} & \multirow{2}{*}{\textbf{\begin{tabular}[c]{@{}l@{}}Max\\ Range\end{tabular}}} \\ \cline{2-7}
		& \multicolumn{1}{l|}{\textbf{P}}     & \multicolumn{1}{l|}{\textbf{T}}   & \multicolumn{1}{l|}{\textbf{S}}   & \multicolumn{1}{l|}{\textbf{AC}}    & \multicolumn{1}{l|}{\textbf{R1}}    & \textbf{AR}  & \multicolumn{1}{l|}{}                                                                                &                                                                               \\ \hline
		GPS                                   & \multicolumn{1}{l|}{1}              & \multicolumn{1}{l|}{1}            & \multicolumn{1}{l|}{1}            & \multicolumn{1}{l|}{1}              & \multicolumn{1}{l|}{1}              & 1            & \multicolumn{1}{l|}{5-50m}                                                                           & 200                                                                           \\
		Gyroscope                             & \multicolumn{1}{l|}{3}              & \multicolumn{1}{l|}{3}            & \multicolumn{1}{l|}{1}            & \multicolumn{1}{l|}{3}              & \multicolumn{1}{l|}{3}              & 3            & \multicolumn{1}{l|}{0.5-9.47 rad}                                                                    & 100                                                                           \\
		Accelerometer                         & \multicolumn{1}{l|}{3}              & \multicolumn{1}{l|}{3}            & \multicolumn{1}{l|}{1}            & \multicolumn{1}{l|}{3}              & \multicolumn{1}{l|}{3}              & 3            & \multicolumn{1}{l|}{0.5-6.2 rad/s2}                                                                  & 26                                                                            \\
		Magnetometer                          & \multicolumn{1}{l|}{3}              & \multicolumn{1}{l|}{3}            & \multicolumn{1}{l|}{1}            & \multicolumn{1}{l|}{1}              & \multicolumn{1}{l|}{3}              & 1            & \multicolumn{1}{l|}{180 deg}                                                                         & -                                                                             \\
		Barometer                             & \multicolumn{1}{l|}{1}              & \multicolumn{1}{l|}{2}            & \multicolumn{1}{l|}{1}            & \multicolumn{1}{l|}{1}              & \multicolumn{1}{l|}{1}              & 1            & \multicolumn{1}{l|}{0.1 kPa}                                                                         & -                                                                             \\ \hline
	\end{tabular}
	\label{tab:attack-recovery-implementation}
\end{table}

\subsection{Comparison with Prior Work} 
\noindent
\textbf{Attack Recovery:}
We quantitatively compare \sysname with three prior attack recovery techniques discussed earlier namely SSR~\cite{srr-choi}, PID-Piper~\cite{pid-piper}, and LQR-O~\cite{recovery-lqr}. 
Recall that LQR-O refers to the recovery technique proposed by Zhang \etal which performs worst-case recovery irrespective of the number of sensors targeted under SDA. 
We collected mission traces of RVs, including both training and testing sets, and assessed the efficacy of existing recovery techniques by reproducing their effectiveness as reported in the respective papers. 
Subsequently, we evaluate all three recovery techniques under SDAs to observe their effectiveness under multi-sensor SDAs targeting different numbers of sensors in the RV.  

\noindent
\textbf{Attack Diagnosis:}
There is no prior work that addresses attack diagnosis in RVs (to the best of our knowledge).  
We extend two existing model-based attack detection techniques namely Savior~\cite{savior} and PID-Piper~\cite{pid-piper}, to serve as a potential solution for attack diagnosis. 
Model-based attack detectors estimate RV's physical states and compare them with the physical states derived from sensors to compute a residue $r = |Model_{states} - Sensor_{states}|$.
Traditionally, these attack detectors analyze residues in just one or two of the RV's physical states, sufficient for flagging an attack.
We extend the concept of residual analysis (RA) to monitor all the physical states in RVs for attack diagnosis. 
Furthermore, RV's are equipped with EKF that estimates the RV's physical states, and uses sensor measurements to enhance state estimations (Section~\ref{sec:bg-control}).  
EKF estimations can be potentially used for attack detection and attack diagnosis by performing residual analysis to monitor EKF estimations and sensor states. 
When a residue for a specific state exceeds the threshold, the corresponding sensor is deemed to be under attack (Table~\ref{tab:states2sensor} shows states to sensor mapping). 
We compare \sysname with the above three baseline RA-based diagnosis techniques.



\subsection{Metrics}

\noindent
\textbf{Recovery Success Metric:}
As in our prior work~\cite{pid-piper}, we consider a mission to be successful, if upon completion, the total deviation from the original destination is {\em less than $10$m.}  
Most GPS sensors used in commodity RVs have an offset of 5 meters~\cite{gps-offset}. 
We consider 2$X$ of the GPS offset as our threshold \ie $5$m offset from the RV's position, and $5$m offset from the destination. 
This 10m threshold is indistinguishable from the standard GPS error \cite{gps-offset}. 
We consider the mission to be unsuccessful if the RV crashes (physically damaged) or stalls (freezes or stops moving towards the destination). 

\smallskip
\noindent
\textbf{Recovery Stability Metric:}
To measure the stability of the RV during recovery we use the Root Mean Square Deviation (RMSD) of the RV's attitude \ie roll, pitch, and yaw angles. 
This metric quantifies the average magnitude of deviations in a recovery-activated mission compared to an attack-free (ground truth) mission on the same trajectory. 
We calculate RMSD using the same method as Root Mean Square Error, as depicted in Equation~\ref{eqn:rmsd}.
As minor attitude deviations between two RV missions even on the same trajectory is expected, we account for these expected deviations by normalizing RMSD with respect to the minimum and maximum expected deviations for an RV on a specific trajectory.
(details in Appendix~\ref{app:metrics}). 
\begin{equation}
\thinspace
    RMSD = \sqrt{\frac{1}{n}\sum_{i=1}^{n}(x_{recovery}-x_{groundTruth})^2}
    \label{eqn:rmsd}
\end{equation}

\noindent
\textbf{Mission Delay due to Recovery:}
We compare the mission completion time between a recovery activated mission with an attack-free ground truth mission on the same trajectory.  
This allows us to calculate potential mission delays resulting from the recovery process.
Given that, there might be minor variations in the mission completion times even in the same trajectory, we normalize the mission delay using baseline mission completion time $T_{baseline}$ for an RV on a given trajectory (details in Appendix~\ref{app:metrics}). 
The percentage mission delay is calculated as shown in Equation~\ref{eqn:pmd}.
\begin{equation}
\thinspace
    PMD = \frac{T_{recovery}-T_{groundTruth}}{T_{baseline}} x 100
    \label{eqn:pmd}
\end{equation}

\noindent
\textbf{Runtime Overheads}: We measure the overheads in CPU times, battery consumption, and memory incurred by the RV due to \sysname. 
As the overheads in simulated RVs depend on the simulation platform, we report the overheads only for the real RVs. 
For CPU overheads, we measure the CPU times incurred by the autopilot modules with and without \sysname. 
During the recovery process, the RV executes the prescribed recovery control actions in place of the actions generated by the PID controller. As a result, the RV's motor rotation rates increase.
We call this operational overhead. 
We estimate the battery overhead of \sysname based on both the CPU overhead and the operational overhead.
Finally, we estimate the memory overhead for storing the historic states on the RV corresponding to all the RV's sensors for targeted attack recovery.

\begin{table*}[!ht]
	\centering
	\scriptsize
	\caption{$\delta$ values for per sensor factor graphs in each subject RVs. position $(x, y, z)$ in meters (m), velocity $(\dot{x}, \dot{y}, \dot{z})$ in $m/s$, acceleration $(\ddot{x}, \ddot{y}, \ddot{z})$ in $m/s^2$, Euler angles $(\phi, \theta, \psi)$ in degrees, angular velocities ($\omega_\phi, \omega_\theta, \omega_\psi$) in rad/s, magnetic fields ($x_m, y_m, z_m$) in Gauss(G), Altitude is in meters. (values rounded up to 1st decimal place). WS: Window size for historic states recording. 
 }
	\begin{tabular}{l|cccccc|ccc|cccccc|ccc|c|c|c|c|c}
		\hline
		\multirow{2}{*}{\textbf{RV Type}} & \multicolumn{6}{c|}{\textbf{GPS}}                                                                                                                                                                                  & \multicolumn{3}{c|}{\textbf{Accel}}                                                                       & \multicolumn{6}{c|}{\textbf{Gyroscope}}                                                                                                                                                                                                  & \multicolumn{3}{c|}{\textbf{Mag}}                                                          & \textbf{Baro} & \multirow{2}{*}{\textbf{\begin{tabular}[c]{@{}c@{}}WS\end{tabular}}} & \multirow{2}{*}{\textbf{CPU}} & \multirow{2}{*}{\textbf{Battery}} & \multirow{2}{*}{\textbf{Memory}} \\ \cline{2-20}
		& \multicolumn{1}{c|}{\textbf{$x$}} & \multicolumn{1}{c|}{\textbf{$y$}} & \multicolumn{1}{c|}{\textbf{$z$}} & \multicolumn{1}{c|}{\textbf{$\dot{x}$}} & \multicolumn{1}{c|}{\textbf{$\dot{y}$}} & \textbf{$\dot{z}$} & \multicolumn{1}{c|}{\textbf{$\ddot{x}$}} & \multicolumn{1}{c|}{\textbf{$\ddot{y}$}} & \textbf{$\ddot{z}$} & \multicolumn{1}{c|}{\textbf{$\phi$}} & \multicolumn{1}{c|}{\textbf{$\theta$}} & \multicolumn{1}{c|}{\textbf{$\psi$}} & \multicolumn{1}{c|}{\textbf{$\dot{\phi}$}} & \multicolumn{1}{c|}{\textbf{$\dot{\theta}$}} & \textbf{$\dot{\psi}$} & \multicolumn{1}{c|}{\textbf{$x_m$}} & \multicolumn{1}{c|}{\textbf{$y_m$}} & \textbf{$z_m$} & \textbf{Alt}  &                                                                                 &                               &                                   &                                  \\ \hline
		Pixhawk                           & \multicolumn{1}{c|}{3.4}          & \multicolumn{1}{c|}{5.1}          & \multicolumn{1}{c|}{5.2}          & \multicolumn{1}{c|}{2.1}                & \multicolumn{1}{c|}{10.2}               & 5.5                & \multicolumn{1}{c|}{6.5}                 & \multicolumn{1}{c|}{4.2}                 & 10.5                & \multicolumn{1}{c|}{12.2}            & \multicolumn{1}{c|}{9.8}               & \multicolumn{1}{c|}{45.5}            & \multicolumn{1}{c|}{11.5}                  & \multicolumn{1}{c|}{13.8}                    & 1.1                   & \multicolumn{1}{c|}{0.6}            & \multicolumn{1}{c|}{0.3}            & 0.5            & 0.2           & 15.5s                                                                           & 8.8\%                         & 22\%                              & 0.47MB                           \\ 
		Tarrot                            & \multicolumn{1}{c|}{5.5}          & \multicolumn{1}{c|}{3.4}          & \multicolumn{1}{c|}{6.5}          & \multicolumn{1}{c|}{1.8}                & \multicolumn{1}{c|}{8.1}                & 4.2                & \multicolumn{1}{c|}{8.1}                 & \multicolumn{1}{c|}{7.7}                 & 9.2                 & \multicolumn{1}{c|}{10.2}            & \multicolumn{1}{c|}{9.5}               & \multicolumn{1}{c|}{38.9}            & \multicolumn{1}{c|}{10.2}                  & \multicolumn{1}{c|}{10.6}                    & 1.1                   & \multicolumn{1}{c|}{0.3}            & \multicolumn{1}{c|}{0.3}            & 0.4            & 0.2           & 15s                                                                             & 6.7\%                         & 18.75\%                           & 0.45MB                           \\ 
		Sky-Viper                         & \multicolumn{1}{c|}{4.6}          & \multicolumn{1}{c|}{3.0}          & \multicolumn{1}{c|}{4.1}          & \multicolumn{1}{c|}{1.7}                & \multicolumn{1}{c|}{7.8}                & 3.1                & \multicolumn{1}{c|}{5.5}                 & \multicolumn{1}{c|}{4.2}                 & 5.6                 & \multicolumn{1}{c|}{13.3}            & \multicolumn{1}{c|}{11.2}              & \multicolumn{1}{c|}{58.5}            & \multicolumn{1}{c|}{14.5}                  & \multicolumn{1}{c|}{16.1}                    & 1.5                   & \multicolumn{1}{c|}{0.6}            & \multicolumn{1}{c|}{0.6}            & 0.5            & 0.2           & 17s                                                                             & 9.2\%                         & 20\%                              & 0.52MB                           \\ 
		AionR1                            & \multicolumn{1}{c|}{2.7}          & \multicolumn{1}{c|}{2.5}          & \multicolumn{1}{c|}{-}            & \multicolumn{1}{c|}{1.9}                & \multicolumn{1}{c|}{6.4}                & -                  & \multicolumn{1}{c|}{3.3}                 & \multicolumn{1}{c|}{3.5}                 & -                   & \multicolumn{1}{c|}{-}               & \multicolumn{1}{c|}{-}                 & \multicolumn{1}{c|}{45.2}            & \multicolumn{1}{c|}{-}                     & \multicolumn{1}{c|}{-}                       & 2.2                   & \multicolumn{1}{c|}{}               & \multicolumn{1}{c|}{}               &                & -             & 18.5s                                                                           & 5.5\%                         & 14.4\%                            & 0.56MB                           \\ 
		ArduCopter                        & \multicolumn{1}{c|}{3.7}          & \multicolumn{1}{c|}{3.3}          & \multicolumn{1}{c|}{4.5}          & \multicolumn{1}{c|}{2.1}                & \multicolumn{1}{c|}{9.3}                & 4.2                & \multicolumn{1}{c|}{5.2}                 & \multicolumn{1}{c|}{6.5}                 & 7.1                 & \multicolumn{1}{c|}{9.7}             & \multicolumn{1}{c|}{9.5}               & \multicolumn{1}{c|}{36.2}            & \multicolumn{1}{c|}{9.5}                   & \multicolumn{1}{c|}{11.2}                    & 1.1                   & \multicolumn{1}{c|}{0.3}            & \multicolumn{1}{c|}{0.2}            & 0.5            & 0.1           & 15.5s                                                                           & -                             & -                                 & -                                \\ 
		ArduRover                         & \multicolumn{1}{c|}{4.2}          & \multicolumn{1}{c|}{4.1}          & \multicolumn{1}{c|}{-}            & \multicolumn{1}{c|}{2.4}                & \multicolumn{1}{c|}{7.3}                & -                  & \multicolumn{1}{c|}{3.5}                 & \multicolumn{1}{c|}{3.9}                 & -                   & \multicolumn{1}{c|}{-}               & \multicolumn{1}{c|}{-}                 & \multicolumn{1}{c|}{38.5}            & \multicolumn{1}{c|}{-}                     & \multicolumn{1}{c|}{-}                       & 1.8                   & \multicolumn{1}{c|}{}               & \multicolumn{1}{c|}{}               &                & -             & 17s                                                                             & -                             & -                                 & -                                \\ \hline
	\end{tabular}
	\label{tab:fg-delta}
\end{table*}

\subsection{Attack Parameters}
\label{sec:implementation}
\noindent
\textbf{Attacks:}
As we did not have access to special equipment (\eg noise emitter, sound source, amplifier, etc.) for mounting physical attacks, we emulated the attacks through targeted software modifications, similar to what most prior work has done \cite{srr-choi, choi, savior, pid-piper}.  
Our attack code interfaces with the sensor libraries in the RV, and manipulates sensor measurements by adding a bias to them (\ie false data).  
When an attack command is initiated, the bias values are automatically added to the raw sensor measurements. 
We derive the attack parameters \ie bias values and attack range to mimic realistic physical attacks via software as closely as possible. 

\smallskip
\noindent
\textbf{Sensor Bias Values:}
We use variable bias values for each sensor within the allowable limit to cause mission delays, instability, and even mission failure.  
We derive bias values as per the respective sensor specifications. 
Table~\ref{tab:attack-recovery-implementation} shows the bias values used for each sensor. 
For example, the update frequency of the GPS module used in many industrial and commodity RVs is 0.1s, and the operational limit in its velocity is 500 m/s \cite{gps-datasheet}. 
Therefore, the maximum hopping distance of the GPS receiver is 50m (update frequency $\times$ maximum velocity). 
Thus, for GPS, we set the bias value to be between 5-50m, which is the operating limit of the GPS sensor.     

\smallskip
\noindent
\textbf{Attack Range:}
We derive the maximum attack range based on prior work that performed the respective attacks through signal injection~\cite{gpsspoofing2, tractor-beam, sensor-confusion}.
We extrapolate their results considering the largest available signal source and amplifier to derive the maximum attack range for each sensor. 
We found that the optimal attack range for individual sensors is within 26 $-$ 200m. 
Of all the sensors, the GPS has the maximum attack range~\cite{gpsspoofing2} of 200m. 
{\em Thus, we select 200m as the attack range for each instance of SDAs in our experiments \ie we assume that attackers can manipulate all or any subset of the RV's sensors in a 200m range.} 
This is a stronger assumption than that made in all the prior work in the area~\cite{gyroscopespoofing, injected-delivered}.

\subsection{\sysname Modules and Parameters}
\textbf{Attack Detector:}
As mentioned, \sysname relies on an attack detection technique, and it provides attack diagnosis and targeted attack recovery after the attack is detected.
For attack detection, we use our prior work, PID-Piper's  attack detection module~\cite{pid-piper-code}, as it represents the state-of-the-art and has a high detection rate.
We do not measure the effectiveness of the attack detector (true positives, and false positives) - this is reported in our prior work~\cite{pid-piper}.  


\smallskip
\noindent
\textbf{Attack Diagnosis:}
The attack diagnosis module monitors the error $e$ between the past and present states of the RV. 
Recall that in the attack-free phase,  $e$ remains within $0 - \delta$. 
This holds true regardless of whether the RV is in a steady or non-steady state.
We use the standard deviation method to derive the value of $\delta$ \ie $\delta = median(e)+ k*stdev(e)$ 
\ie if the error $e$ is more than $k$ standard deviations away from the mean, it is likely to be an outlier (due to the attack)~\cite{threshold-method}.
We collect attack-free RV mission traces (on both simulated and real RVs with diverse mission trajectories), and empirically determine that $k=3$ ensures $0<e<\delta$ in the attack-free phase of the mission. 
We run between 15-25 attack-free missions for each RV to derive the $\delta$ values, and we validated $\delta$ values (\ie $e$ $\in$ $0-\delta$) by running another 15 missions. 

\begin{figure}[!ht]
	\centering
	\begin{subfigure}{.49\linewidth}
		\centering
		\includegraphics[width=\linewidth]{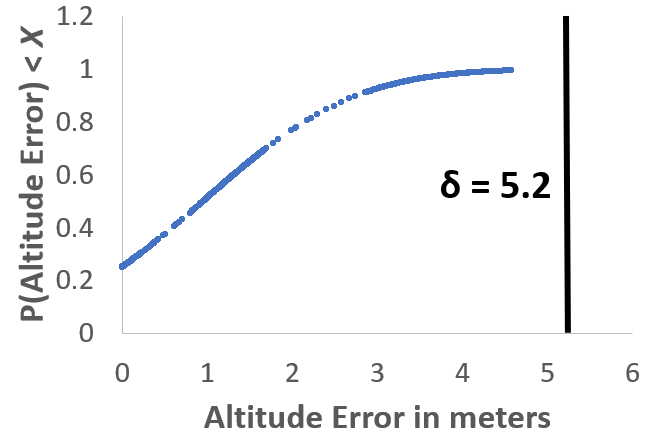}
		\caption{}
    	\label{fig:delta-value}
	\end{subfigure}
	\begin{subfigure}{.49\linewidth}
		\centering
		\includegraphics[width=\linewidth]{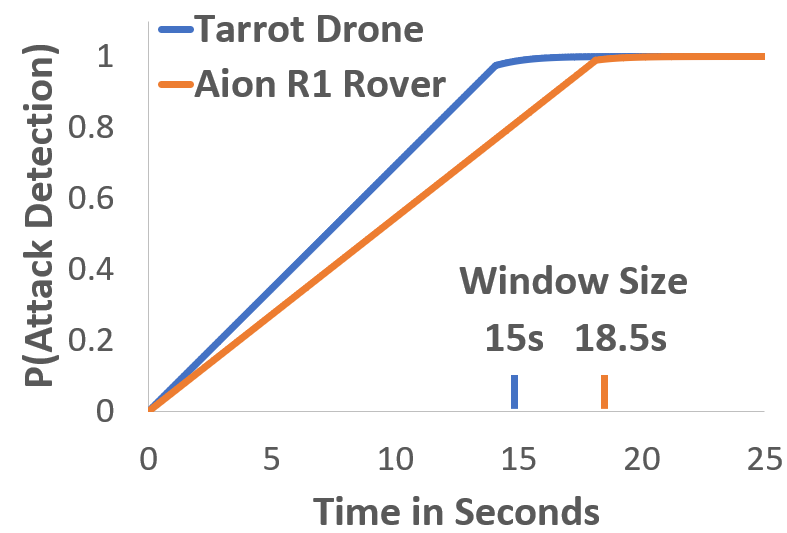}
		\caption{}
        \label{fig:window-cdf}
	\end{subfigure}
	\caption{ 
        (a) CDF of error distribution in the Pixhawk drone. $k=3$ ensures $0<e<\delta$ in attack-free phase. 
        (b) CDF of the likelihood of stealthy attack detection. 
        Window size of 15s (Tarrot) and 18.5s (Aion A1) ensures 100\% attack detection.  
        } 
\end{figure}

Figure~\ref{fig:delta-value} shows an example of $\delta$ for position error in the Pixhawk drone.  
The figure shows the Cumulative Distribution Function (CDF) of error $e$ as a function of the probability that the physical states errors will take a value less than $\delta$. 
With k=3, we obtain the $\delta$ value for $z$-axis position error as $5.2$.
As shown in Figure~\ref{fig:delta-value}, in the attack-free phase, the error $e$ always remains under the $\delta$.
Table~\ref{tab:fg-delta} shows the $\delta$ values for all the physical states of our subject RVs. 

\smallskip
\noindent
\textbf{Attack Recovery under Stealthy Attacks:}
Recall that we prevent disruptions due to stealthy attacks by setting the window size for recording historic states to be large enough that stealthy attacks can be detected in a single sliding window (Section~\ref{sec:hsr}).
The key question is what is the maximum duration that the attacker can cause disruptions in the RV mission by launching stealthy attacks while evading detection. 
Subsequently, we can set an appropriate window size for the effective detection of stealthy attacks.
Therefore, to find the appropriate window size, we launched two kinds of stealthy attacks by injecting persistent bias: 
(1) targeting all the sensors simultaneously, 
(2) targeting each sensor individually.  

Note that the above stealthy attacks were launched on all of our subject RVs, in the presence of PID-Piper's attack detector~\cite{pid-piper} that uses CUSUM to detect stealthy attacks~\cite{stealthy-attacks}. 
We find that when all the sensors are manipulated persistently, the stealthy attack is detected in under 3.3s. 
However, when the attack targets individual sensors, the stealthy attacks against GPS remain undetected for the maximum duration \ie around $15-18s$.

We determine the window size for each RV to be larger than the duration of a stealthy attack that targets GPS. 
Figure~\ref{fig:window-cdf} shows a CDF that depicts for how long a stealthy attack targeting GPS can evade detection as the mission progresses. 
100\% attack detection rate is achieved for Tarrot drone at 15s, and for Aion R1 rover at 18.5s. 
{Thus, we select window sizes of 15s and 18.5s for the respective RVs, as the likelihood of detecting stealthy attacks in that window is near 100\%.}
Table~\ref{tab:attack-recovery-implementation} shows the window sizes for all RVs derived in the same manner.

\section{Results}
\label{sec:results}
We first evaluate \sysname's attack diagnosis under SDAs in a simulated RV. 
Then, we evaluate \sysname's targeted attack recovery and compare it with other techniques under SDAs in the simulated RVs. 
We then study the significance of attack diagnosis under multi-sensor SDAs, followed by \sysname's attack diagnosis and recovery in real RVs. 
We also study \sysname's effectiveness under stealthy attacks. Finally, we  present the overheads incurred by \sysname. 

\subsection{Effectiveness of \sysname's Attack Diagnosis}
\label{sec:results-attack-diagnosis}

In this section, we evaluate the effectiveness of \sysname's attack diagnosis by measuring the true positives (TP) and false positives (FP). 
TP is the fraction of times attack diagnosis correctly identified the sensors targeted by SDA. 
FP is the fraction of times attack diagnosis falsely identified sensors as attacked even in the absence of an attack.
because no prior work has addressed attack diagnosis, we extended the residual analysis in existing attack detection techniques namely Savior~\cite{savior}, PID-Piper~\cite{pid-piper}, and EKF~\cite{ekf}, for attack diagnosis. 
We compare \sysname's diagnosis that uses factor graphs with techniques that use residual analysis (RA-based diagnosis). 

We run 100 missions using simulated RVs and launched SDAs targeting up to $n-1$ sensors of the RV. 
We launched the same attacks for all the diagnosis techniques to compare them.  
Table~\ref{tab:attack-diagnosis} shows the results.
We find that, Savior, PID-Piper, and EKF based diagnosis techniques achieve on average 60\%, 56\%, and 62.5\% TPs respectively. 
In contrast, \sysname's diagnosis correctly identified the targeted sensors in 95\% of the cases (on average). 
This is because \sysname attributes manipulations in the RV's physical states to a sensor being under attack, by performing causal analysis using factor graphs. 
In contrast, the RA-based techniques perform a naive attack diagnosis by simply comparing the residuals against a threshold, thus failing to achieve a high diagnosis TP rate. 
{\em Thus, \sysname's diagnosis resulted in 40 to 50\% higher TPs rate on average than RA-based diagnosis methods (Table~\ref{tab:attack-diagnosis})}. 
When SDAs targeting all the sensors were launched, all the techniques including \sysname achieve a high diagnosis TP rate. 

\smallskip
\noindent
\textbf{False Positives and Gratuitous Recovery Activation.}
Note that a false alarm in detection leads to an FP in attack diagnosis, and eventually results in gratuitous recovery activation.
During gratuitous recovery, the attack recovery technique generates unnecessary recovery control actions, causing needless maneuvers that result in mission disruptions and delays.
We evaluate \sysname's FPs in the absence of attacks to assess its gratuitous recovery activations. 

Attack detection techniques raise false alarms when exposed to environmental disturbances such as wind. 
To measure FPs in attack diagnosis, we induce false alarms in attack detectors by simulating wind conditions. 
It is important to strike a balance between the attack detection accuracy and induced false alarms, and not overwhelm the attack detector with an excessive number of false alarms. 
Thus, we simulated a modest wind speed of 15 km/h~\cite{wind-scale}.

\begin{table}[!ht]
\footnotesize
\caption{\sysname's attack diagnosis compared with residual analysis-based diagnosis of Savior, PID-Piper, and EKF. Values are in Percentage.}
\begin{tabular}{c|c|c|c|c}
\hline
\textbf{\begin{tabular}[c]{@{}c@{}}Diagnosis TPs \\ number of sensors\\ targeted by SDA\end{tabular}} & \textbf{\begin{tabular}[c]{@{}c@{}}Savior \\ RA\end{tabular}} & \textbf{\begin{tabular}[c]{@{}c@{}}PID-Piper \\ RA\end{tabular}} & \textbf{\begin{tabular}[c]{@{}c@{}}EKF \\ RA\end{tabular}} & \textbf{\sysname} \\ \hline
1                                                                                                & 65                                                            & 65                                                               & 75                                                         & 100               \\ 
2                                                                                                & 55                                                            & 50                                                               & 60                                                         & 100               \\ 
3                                                                                                & 50                                                            & 50                                                               & 55                                                         & 90                \\ 
4                                                                                                & 70                                                            & 60                                                               & 60                                                         & 90                \\ 
Average                                                                                          & 60                                                            & 56                                                               & 62.5                                                       & 95                \\ \hline
FPs (no attack)          & 20                                                            & 15                                                               & 20                                                         & 5                 \\ \hline
\end{tabular}
\label{tab:attack-diagnosis}
\end{table}

We run 20 missions in Arducopter for each technique, activating diagnosis when attack detectors raised false alarms. Table~\ref{tab:attack-diagnosis} shows the FPs in each technique. 
The corresponding attack detectors of each RA-based diagnosis technique - Savior, PID-Piper, and EKF reported false alarms in 20\%, 15\%, and 20\% of the cases respectively.
As RA-based diagnosis methods rely on attack detectors’ state estimations they also incurred 15-20\% FPs. 
In contrast, DeLorean masked 15\% false alarms from the attack detector and incurred only 5\% FPs i.e., 4X lower FPs than other diagnosis techniques.

The RA-based diagnosis techniques resulted in significantly higher FPs because they monitor the physical states' error only in the last step, and hence fail to account for the benign deviations in RV's physical states.  
In contrast, \sysname's diagnosis monitors the sequence of errors in four previous steps (this is inherent to the FG approach, see Figure~\ref{fig:factor-graphs}).
Thus, FG captures the causality of manipulations in RV's physical states to infer if the corresponding sensor is under attack, and achieves both high TPs and low FPs.
Furthermore, RA-based diagnosis relies on the attack detector's response, false alarms in detection propagate as FPs in diagnosis. 
\sysname performs attack diagnosis independently, allowing it to mask FPs triggered during attack detection.

The FPs in attack diagnosis resulted in gratuitous recovery activation.
We find that the gratuitous recovery operations due to FPs in RA-based diagnosis resulted in RMSD of 0.095 and mission delays of more than 10\% (on average). 
In contrast \sysname incurred almost negligible RMSD of 0.0045 which resulted in $\approx1\%$ mission delays.  
{\em Thus, \sysname achieves a significant reduction of 4X in gratuitous recovery activation and prevents unnecessary mission disruptions.}   

\subsection{Effectiveness of \sysname's Attack Recovery}
\label{sec:attack-recovery-results}
We compare \sysname with SSR, PID-Piper, and \lqr in recovering from the SDAs on the simulated RVs.
We launched 100 missions for each simulated RV protected with SSR, PID-Piper, \lqr, and \sysname respectively. 
We mounted the same SDAs for all four techniques varying the number of sensor types targeted from 1 up to 5 (all sensors). 
Note that when we say a sensor is attacked, we mean that {\em all} the sensors of that type are attacked in the SDA.

\begin{table}[!ht]
\centering
\footnotesize
\caption{\sysname's recovery outcomes compared with SSR, PID-Piper, and \lqr as a function of the number of sensors attacked. MS: Mission Success, values are in percentages.}
\begin{tabular}{c|cc|cc|cc|cc}
\hline
\multirow{2}{*}{\textbf{\begin{tabular}[c]{@{}c@{}}\# of sensors \\ targeted\end{tabular}}} & \multicolumn{2}{c|}{\textbf{SSR}}                 & \multicolumn{2}{c|}{\textbf{PID-Piper}}           & \multicolumn{2}{c|}{\textbf{LQR-O}}               & \multicolumn{2}{c}{\textbf{DeLorean}}            \\ \cline{2-9} 
                                                                                            & \multicolumn{1}{c|}{\textbf{Crash}} & \textbf{MS} & \multicolumn{1}{c|}{\textbf{Crash}} & \textbf{MS} & \multicolumn{1}{c|}{\textbf{Crash}} & \textbf{MS} & \multicolumn{1}{c|}{\textbf{Crash}} & \textbf{MS} \\ \hline
1                                                                                           & \multicolumn{1}{c|}{20}             & 64          & \multicolumn{1}{c|}{0}              & 100         & \multicolumn{1}{c|}{0}              & 90          & \multicolumn{1}{c|}{0}              & 100         \\ 
2                                                                                           & \multicolumn{1}{c|}{56}             & 20          & \multicolumn{1}{c|}{64}             & 20          & \multicolumn{1}{c|}{0}              & 84          & \multicolumn{1}{c|}{0}              & 100         \\ 
3                                                                                           & \multicolumn{1}{c|}{100}            & 0           & \multicolumn{1}{c|}{100}            & 0           & \multicolumn{1}{c|}{0}              & 84          & \multicolumn{1}{c|}{0}              & 100         \\ 
4                                                                                           & \multicolumn{1}{c|}{100}            & 0           & \multicolumn{1}{c|}{100}            & 0           & \multicolumn{1}{c|}{4}              & 82          & \multicolumn{1}{c|}{4}              & 88          \\ 
5                                                                                           & \multicolumn{1}{c|}{100}            & 0           & \multicolumn{1}{c|}{100}            & 0           & \multicolumn{1}{c|}{4}              & 82          & \multicolumn{1}{c|}{4}              & 82          \\ \hline
\end{tabular}
\label{tab:recovery-comparison}
\end{table}

Table~\ref{tab:recovery-comparison} shows the mission outcomes of \sysname, SSR, PID-Piper, and \lqr under single and multi-sensor SDAs. 
From the table, SSR has a mission success rate of 64\% for single sensor attacks. 
However, for attacks targeting two sensors, its mission success rate drops to 20\% with a crash rate of 56\%. 
PID-Piper, on the other hand, is 100\% effective when only a single sensor is targeted. However, its mission success rate also drops to 20\%, and its crash rate increases to 64\% when two sensors are attacked. 
Furthermore, when 3 or more sensors are attacked, both SSR and PID-Piper incurred a 100\% crash rate (0\% success). 
Thus, {\em \mbr recovery techniques SSR and PID-Piper incur crashes and mission failures} for multi-sensor SDAs, and had average mission success rates of 17\% and 24\% respectively. 

On the other hand, \lqr has an average mission success rate of 86\% with no crashes when 1 to 3 sensors are under attack. 
However, when 4 or more sensors are under attack, \lqr's average recovery success drops to 82\%. 
Because we inject variable sensor bias within the allowable range, 
When 4 or more sensors are targeted simultaneously, the SDA triggers aggressive deviations from the intended path influenced by the injected bias values. 
This makes it challenging for \lqr to derive safe recovery control actions. 

In contrast, \sysname ~{\em achieves 100\%} mission success rate when fewer than 3 sensors are attacked, and {\em no crashes}. 
When 4 or more sensors are targeted, \sysname also achieves $85\%$ mission success and 4\% crash rate (same as \lqr).
\sysname achieves {\em $94\%$ mission success rate on average across all sensor numbers, which is $\approx4X$ higher than the mission success rates of SSR, and PID-Piper, and 10\% point increase compared to \lqr.} 	 
{\em Thus, \sysname is more resilient to multi-sensor SDAs than prior techniques.}

\subsection{Need for Attack Diagnosis}
To understand if attack diagnosis is needed for recovery from SDAs, we compare \sysname's diagnosis guided targeted attack recovery with \lqr under SDAs targeting various numbers of sensors.
Note that \lqr assumes a worst-case recovery scenario and applies recovery operations corresponding to all the sensors irrespective of the number of sensors targeted by an SDA. 

In this experiment, we perform 100 missions each for \sysname and \lqr in simulated RVs, and launch SDAs targeting 1 to 5 sensors in each mission (all combinations). 
In addition to mission success rate and crash rate, we measure Root Mean Squared Deviation (RMSD), and Mission Delay (MD). 
RMSD provides a measure of the stability of the recovery process, and MD shows the impact of recovery on the expected mission duration. 
Lower RMSD and MD signifies the recovery has insignificant impact on the RV mission. 

\begin{table}[!ht]
\footnotesize
\caption{Comparison between \sysname and \lqr (without diagnosis) under SDAs. RMSD: Root Mean Squared Deviation, MD: Mission Delay (shown as the percentage increase in mission duration), MS: Mission Success Rate.}
\begin{tabular}{c|cccc|cccc}
\hline
\multirow{2}{*}{\textbf{\begin{tabular}[c]{@{}c@{}}sensors\\ targeted\end{tabular}}} & \multicolumn{4}{c|}{\textbf{\lqr}}                                                                          & \multicolumn{4}{c}{\textbf{\sysname}}                                                                      \\ \cline{2-9} 
                                                                                     & \multicolumn{1}{c|}{\textbf{RMSD}} & \multicolumn{1}{c|}{\textbf{MD}} & \multicolumn{1}{c|}{\textbf{Crash}} & \textbf{MS} & \multicolumn{1}{c|}{\textbf{RMSD}} & \multicolumn{1}{c|}{\textbf{MD}} & \multicolumn{1}{c|}{\textbf{Crash}} & \textbf{MS} \\ \hline
1                                                                                    & \multicolumn{1}{c|}{0.6259}         & \multicolumn{1}{c|}{14.5}        & \multicolumn{1}{c|}{0}              & 90          & \multicolumn{1}{c|}{0.1236}         & \multicolumn{1}{c|}{2.41}        & \multicolumn{1}{c|}{0}              & 100         \\ 
2                                                                                    & \multicolumn{1}{c|}{0.6319}         & \multicolumn{1}{c|}{14.66}       & \multicolumn{1}{c|}{0}              & 84          & \multicolumn{1}{c|}{0.1552}         & \multicolumn{1}{c|}{3.66}        & \multicolumn{1}{c|}{0}              & 100         \\ 
3                                                                                    & \multicolumn{1}{c|}{0.6414}         & \multicolumn{1}{c|}{14.46}       & \multicolumn{1}{c|}{0}              & 84          & \multicolumn{1}{c|}{0.2733}         & \multicolumn{1}{c|}{7.75}        & \multicolumn{1}{c|}{0}              & 100         \\ 
4                                                                                    & \multicolumn{1}{c|}{0.6383}         & \multicolumn{1}{c|}{12.41}       & \multicolumn{1}{c|}{4}              & 82          & \multicolumn{1}{c|}{0.5784}         & \multicolumn{1}{c|}{10.31}       & \multicolumn{1}{c|}{4}              & 88          \\ 
5                                                                                    & \multicolumn{1}{c|}{0.6603}         & \multicolumn{1}{c|}{12.58}       & \multicolumn{1}{c|}{4}              & 82          & \multicolumn{1}{c|}{0.6603}         & \multicolumn{1}{c|}{12.58}       & \multicolumn{1}{c|}{4}              & 82          \\ \hline
\end{tabular}
\label{tab:why-diagnosis}
\end{table}

Table~\ref{tab:why-diagnosis} shows the comparison between \sysname and \lqr. 
As we have already discussed the mission success and crash rates for both techniques in Section~\ref{sec:attack-recovery-results}, we do not repeat them here. 
We find that the RMSD values of \lqr do not vary much across SDAs that target varying numbers of sensors. 
In comparison, \sysname has 2.2X lower RMSD (on average) than \lqr when SDAs target a subset of the RV's sensors.   
Furthermore, when up to 3 sensors are targeted, \sysname's recovery incurs 2X to 5X less RMSD compared to \lqr. 
This is because \lqr applies the worst-case recovery strategy in all cases. 
Such recovery strategy results in \lqr applying aggressive maneuvers when a subset of sensors is targeted. 
Unsurprisingly, both \sysname and \lqr have the same RMSD values when all the sensors are targeted as both techniques perform the same recovery operations in such a situation. 
{Thus, \em attack diagnosis-guided attack recovery as done in \sysname ensures minimal deviations under SDA targeting any combination of sensors.} 

Furthermore, we find that \lqr recovery on average, incurs ~2.5X higher mission delay than \sysname when the SDA targets only a subset of the sensors. 
Recall that \lqr recovery results in higher deviations from the trajectory compared to \sysname, consequently \lqr requires more time to restore the RV to its intended path. 
Further, as shown in Table~\ref{tab:why-diagnosis}, \lqr's mission delays are consistent for attacks targeting varying numbers of sensors. 
On the other hand,  \sysname's mission delays for attacks targeting less than 3 sensors are significantly lower than that of \lqr. 
When a single sensor is targeted \lqr's recovery increases the expected mission duration by 14.5\%. 
In contrast, \sysname increases the mission duration only by 2.4\%. 
However, for attacks targeting 4 or 5 sensors simultaneously \sysname incurs similar mission delays as \lqr. 
{\em Thus, \sysname's attack diagnosis guided attack recovery prevents unnecessary mission delays while recovering from multi-sensor SDAs}. 

\subsection{Attack diagnosis and Recovery in Real RVs}
We also evaluate \sysname on the four real RVs under SDAs targeting different numbers of sensors. 
Table~\ref{tab:real-rv-attack-recovery} shows \sysname's diagnosis and recovery outcomes.
We find that \sysname's diagnosis achieves 100\% TP rate when the SDA targets up to 2 sensors simultaneously. 
However, its TP rate drops to between 80\% and 90\% when the SDA targets 3 or more sensors. 
On average, \sysname's attack diagnosis TP rate is between 92\% to 98\% across all RVs. 
We also find that \sysname's diagnosis FP rate was between 2\% and 6\% across all RVs (not presented due to space constraints).
These results are comparable to those on the simulated RVs (Section~\ref{sec:results-attack-diagnosis}).

\begin{table}[!ht]
\footnotesize
\caption{\sysname's attack diagnosis and attack recovery in real RVs. All values shown in percentage.}
\begin{tabular}{c|cc|cc|cc|cc}
\hline
\multirow{2}{*}{\textbf{\begin{tabular}[c]{@{}c@{}}\# of sensors\\ targeted\end{tabular}}} & \multicolumn{2}{c|}{\textbf{Pixhawk}}          & \multicolumn{2}{c|}{\textbf{Tarrot}}           & \multicolumn{2}{c|}{\textbf{Sky-viper}}        & \multicolumn{2}{c}{\textbf{Aion R1}}          \\ \cline{2-9} 
                                                                                           & \multicolumn{1}{c|}{\textbf{TP}} & \textbf{MS} & \multicolumn{1}{c|}{\textbf{TP}} & \textbf{MS} & \multicolumn{1}{c|}{\textbf{TP}} & \textbf{MS} & \multicolumn{1}{c|}{\textbf{TP}} & \textbf{MS} \\ \hline
1                                                                                          & \multicolumn{1}{c}{100}         & 100         & \multicolumn{1}{c|}{100}         & 100         & \multicolumn{1}{c|}{100}         & 100         & \multicolumn{1}{c|}{100}         & 100         \\ 
2                                                                                          & \multicolumn{1}{c|}{100}         & 100         & \multicolumn{1}{c|}{100}         & 100         & \multicolumn{1}{c|}{100}         & 100         & \multicolumn{1}{c|}{100}         & 100         \\ 
3                                                                                          & \multicolumn{1}{c|}{90}          & 100         & \multicolumn{1}{c|}{100}         & 100         & \multicolumn{1}{c|}{80}          & 100         & \multicolumn{1}{c|}{80}          & 100         \\ 
4                                                                                          & \multicolumn{1}{c|}{90}          & 80          & \multicolumn{1}{c|}{90}          & 90          & \multicolumn{1}{c|}{80}          & 80          & \multicolumn{1}{c|}{90}          & 80          \\ 
5                                                                                          & \multicolumn{1}{c|}{100}         & 80          & \multicolumn{1}{c|}{100}         & 80          & \multicolumn{1}{c|}{100}         & 80          & \multicolumn{1}{c|}{100}         & 80          \\ 
Average                                                                                    & \multicolumn{1}{c|}{96}          & 92          & \multicolumn{1}{c|}{98}          & 94          & \multicolumn{1}{c|}{92}          & 92          & \multicolumn{1}{c|}{94}          & 92          \\ \hline
\end{tabular}
\label{tab:real-rv-attack-recovery}
\end{table}

We find that \sysname's average mission success rate for SDAs targeting upto $n-1$ of the RV's sensors is 95.62\%. Furthermore, \sysname successfully recovered RVs in 100\% of the cases when the SDA targets up to 3 sensors simultaneously. 
However, its mission success rate drops to 80\% when SDAs target all the 5 sensors simultaneously. 
{\em \sysname achieves 92.5\% average mission success rate across RVs}. 
Furthermore, we observed no crashes in any of the real RV mission.  
Finally, even for the failed missions, the deviation from the target was very small ($15.6m$ on average).

\begin{figure}[!ht]
\centering
\includegraphics[width=0.37\textwidth]{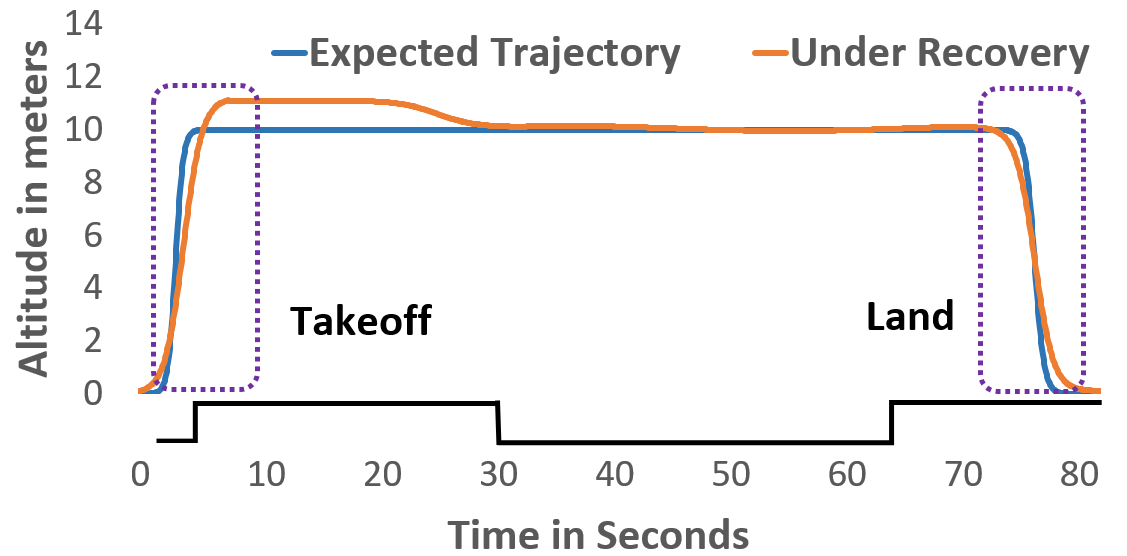}	
\caption{
    \sysname's recovery in the Pixhawk drone under SDAs. 
    Bottom graph shows when the SDA is launched. 
}
\label{fig:pixhawk-recovery}
\end{figure}

Figure~\ref{fig:pixhawk-recovery} shows \sysname's recovery on the real Pixhawk drone when SDAs targeted GPS and accelerometer sensors simultaneously. 
This experiment closely resembles the one discussed in Section~\ref{sec:motivation}. 
We deliberately target 2 out of 5 sensors to compare with \lqr's recovery discussed in Section\ref{sec:motivation} and \sysname's targeted recovery. 
In this mission, the drone should attain an elevation of 10m and navigate in a straight line for 70s, and land when it reaches the destination. 
Two instances of SDAs were launched: (i) during the takeoff from 5s to 30s, (ii) during landing from 65s to 85s.
Note that, in a similar mission, shown earlier in Figure~\ref{fig:motivation}, \lqr incurred an RMSE of 20.66, a mission delay of 18.75\%, and resulted in a mission failure with the drone landing 15m off the target destination.

In contrast, when \sysname is deployed in the Pixhawk drone, it achieved RMSD of 4.21, caused only 2.5\% mission delay, and it completed the mission successfully landing the drone at the destination. 
As shown in Figure~\ref{fig:pixhawk-recovery}, \sysname limited deviations from the expected trajectory to 1.5m under the first SDA during takeoff. 
\sysname maneuvered the drone without incurring any additional delays and landed it safely under the second SDA.
Therefore, \sysname's diagnosis and targeted recovery successfully navigated the drone toward its set target and completed the mission successfully.


\subsection{Recovery under Stealthy Attacks}
Recall that \sysname sets a historic states recording window large enough that a stealthy attack that injects persistent bias can be detected in one sliding window using CUSUM~\cite{pid-piper}. 
An attacker who is aware of this design can launch adaptive stealthy attacks by introducing variations in the bias. 
The attacker's goal is to prolong the stealthy attack duration while evading detection, corrupt the \sysname's recorded historic states, and consequently, affect \sysname's recovery process.
We perform experiments on ArduCopter to evaluate \sysname under three adaptive stealthy attacks. 
Instead of persistent bias, these adaptive attacks inject: (A1) random bias, (A2) gradually increasing bias, and (A3) intermittent bias. 

\begin{figure}[h]
	\centering
	\begin{subfigure}{.235\textwidth}
		\centering
		\includegraphics[width=\linewidth]{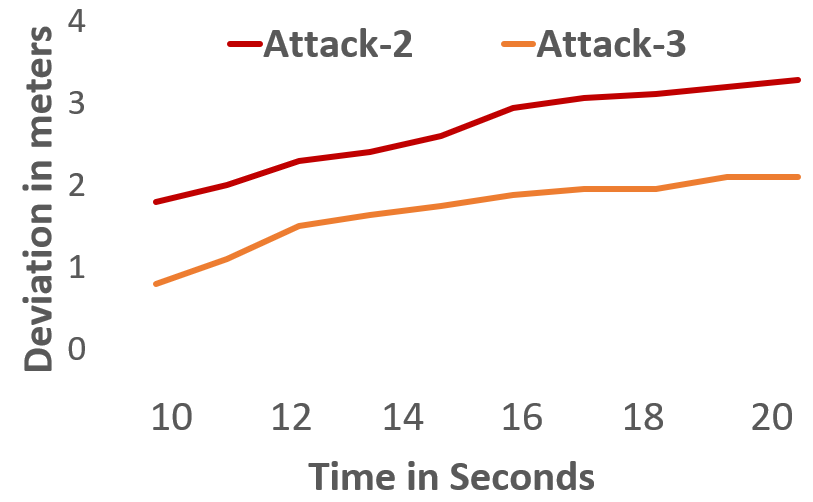}
		\caption{}
		\label{fig:deviation-aa}
	\end{subfigure}
	\begin{subfigure}{0.235\textwidth}
		\includegraphics[width=\linewidth]{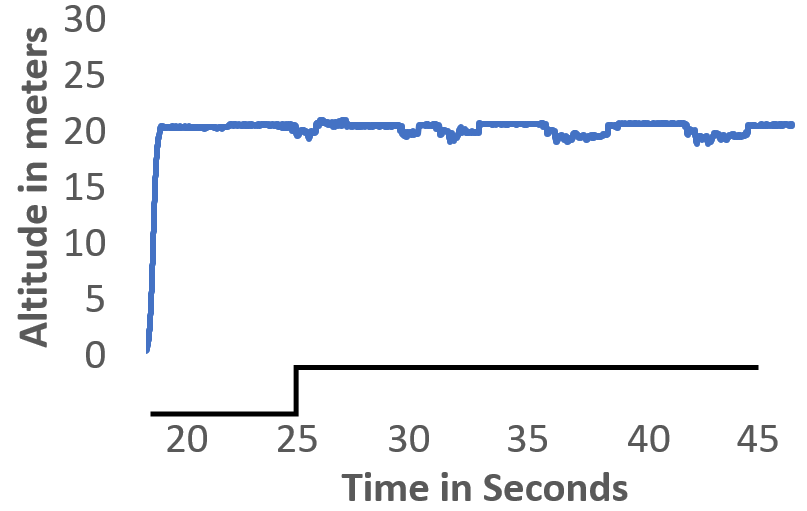}
		\caption{}
		\label{fig:altitude-aa}
	\end{subfigure}
	\caption{
		\sysname recovery in ArduCopter under stealthy attacks.  
		(a) Corruptions in recorded states due to stealthy attacks, (b) Recovery under stealthy attacks.}
	\label{fig:adaptive-attacker}
\end{figure}

In the first experiment, we launched attack A1 following the drone takeoff, and we targeted all the sensors simultaneously. 
We injected random bias causing deviations between 0-5m in the RV's trajectory.  
We found that this attack was detected within one sliding window. 
Thus, attack-A1 failed to corrupt the \sysname's historic states, and  
consequently, \sysname performed a successful recovery. 

In the second experiment, we launched attack A2 starting at t=10s. 
This attack remained undetected in the initial sliding window and caused a total deviation of $3.28m$ from the drone's original path (Figure~\ref{fig:deviation-aa}). 
Thus, attack A2 successfully corrupted \sysname's historic states. 
As no alarm was raised by the attack detector, \sysname recorded corrupted historic states. 
At t=25s the attack was detected and recovery activated. 
As shown in Figure~\ref{fig:altitude-aa}, \sysname was able to successfully recover the drone, maintaining its altitude at 20m despite the corruption in the historic states.   
This is because the corruption in historic states due to attack A2 is negligible, and hence it does not disrupt \sysname's recovery control actions. 

Finally, we evaluated \sysname under attack A3. 
We launched the attack intermittently following the RV's takeoff. 
This attack also managed to evade detection but resulted in a $2.1m$ total deviation from the RV's original path. 
Consequently, \sysname recorded the corrupted historic states similar to the A2 scenario. 
We find that \sysname successfully recovered the RV as the historic states corruptions exhibited negligible impact on the recovery process. 

{\em Thus, \sysname ensures timely detection of a stealthy attack within one sliding window, thereby ensuring no corruption of the recovery process. 
Furthermore, \sysname's robustness to the historic states' corruptions under adaptive stealthy attacks ensures successful recovery.} 

\subsection{Overheads}
Table~\ref{tab:fg-delta} shows the overheads incurred by \sysname on the real RVs. 
The CPU overhead varies from $5.5\%$ to $9.2\%$, with an average of $7.5\%$ across the four RVs.
This is slightly higher than the CPU overhead incurred by SSR and PID-Piper, which are 6.9\% and 6.35\% respectively (on similar RV platforms).
However, because \lqr does not report CPU overhead, we estimate its overhead based on \sysname's overhead. As \lqr does not perform any diagnosis unlike \sysname, we subtract the cost of diagnosis from \sysname's overhead, which comes to an average $7\%$ across the RVs. 

Furthermore, \sysname's memory overhead is between 0.45 MB and 0.56 MB across the different RVs. This accounts for less than 3\% overhead, as the memory available on the real RVs is over 20 MB.

Finally, under attacks, \sysname incurs battery overheads between 14.4\% and 22\% across RVs (18.87\% on average). 
This overhead is a result of recovery control actions and mission delays, causing the motors to remain operational for an extended period.
In comparison, \lqr incurs significantly higher battery overhead as it incurs 2.5X higher mission delays than \sysname.  Note that we did not measure the battery overheads for PID-Piper and SSR as they do not recover from most SDAs. 

\section{Conclusion}
\label{sec:conclusion}
We presented \sysname, a unified framework for attack detection, diagnosis, and recovery from multi-sensor SDAs in RVs.
We proposed a factor graphs based attack diagnosis method that inspects the attack induced errors in the RV's physical states and performs causal analysis to identify the sensors targeted by SDAs.
\sysname then isolates those sensors from RV's feedback control loop to prevent erroneous control actions. 
\sysname integrates existing attack detection and recovery techniques with our proposed attack diagnosis technique to provide targeted attack recovery in RVs. 
We evaluate \sysname on four real and two simulated RVs. 
We find that \sysname (1) recovers the RVs from SDAs, and achieves
mission success in 93\% of the cases, 
(2) minimizes mission disruptions during recovery by performing diagnosis guided targeted recovery, and 
(3) achieves mission success even against stealthy attacks, and (4) incurs  modest performance, memory, and battery overheads.

\section*{Acknowledgements}
This work was partially supported by the Natural Sciences
and Engineering Research Council of Canada (NSERC), and
a Four Year Fellowship from UBC. 
We thank the reviewers of AsiaCCS’24 and our shepherd Dr. Jairo Giraldo for their helpful comments.

\bibliographystyle{ACM-Reference-Format}
\bibliography{main}

\vspace{5mm}
\appendix
\section{Research Methods}
\subsection{Joint Probability and Factor Graphs}
\label{app:factor-graphs}
The joint probability of variables $x_1, ..., x_n$ is described as $P(x_1,..,x_n)$. 
Causal inference of an event given the observed variable is calculated as: 
$y = argmax P(y|x_1,..,x_{n-1})$, which can be further expanded as the following.  
\begin{equation}
	P(y|x_1,..., x_{n-1}) = \frac{P(x_1,..,x_{n-1})}{\sum_v P(x_1,.., x_{n-1}, y=v)}
\end{equation}

Inference on a joint probability distribution $P(x_1,..,x_n)$ requires $2^n$ storage for events with a binary outcome (\eg malicious, benign), which is computationally expensive as well.

\begin{figure}[!ht]
	\centering
	\includegraphics[width=0.15\textwidth]{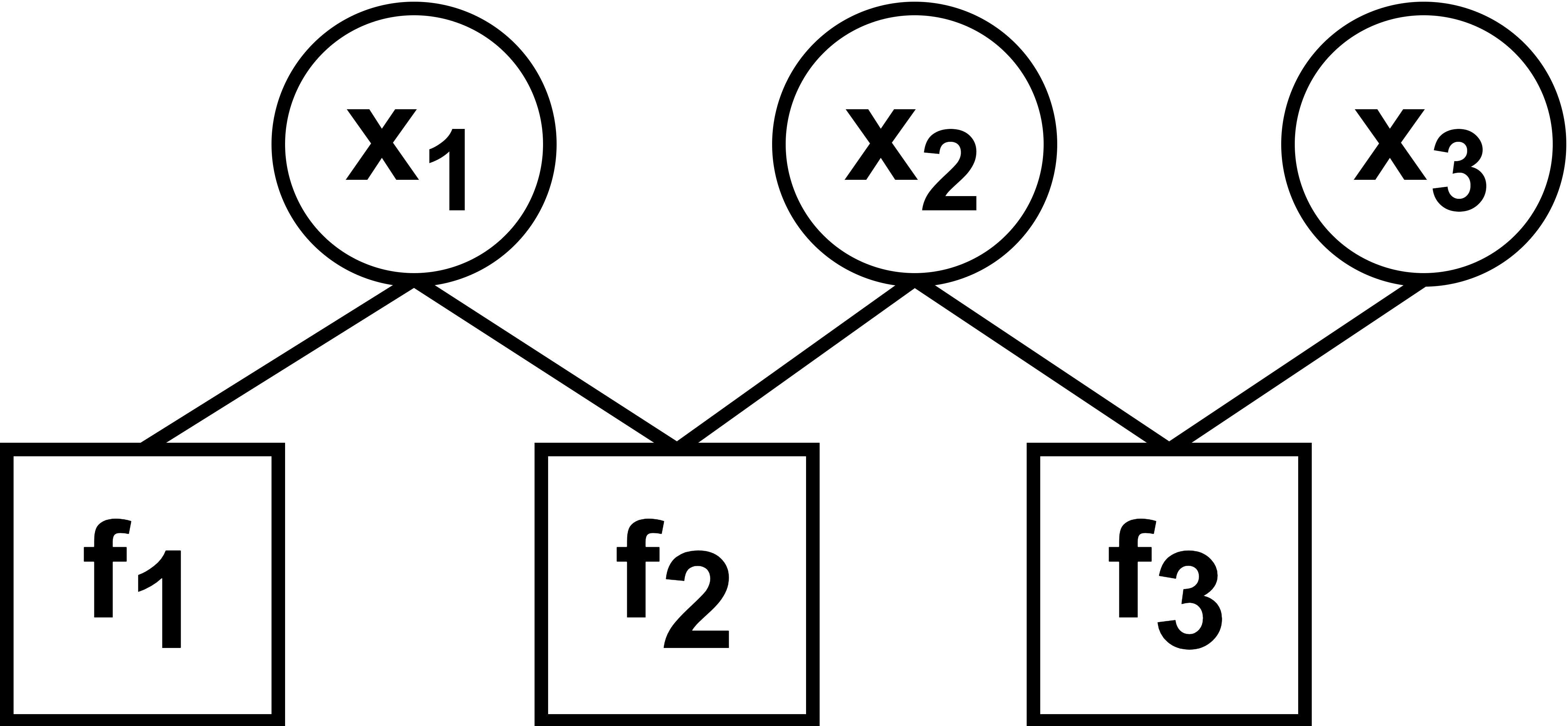}
	\caption{An example of factor graph}
	\label{fig:fg-example}
\end{figure}

Factor graphs can be used to overcome this problem. 
A factor graph (FG) is a probabilistic graphical model that allows expressing joint probability as a product of smaller local functions~\cite{factor-graph}.
Figure~\ref{fig:fg-example} shows an example. 
There are two types of nodes in a FG namely variables ($x_1, x_2, x_3$) and factor functions ($f_1, f_2, f_3$).
Variables are used to quantitatively describe an event.
A factor function is used to express relationships among variables. 

Let's take Figure~\ref{fig:fg-example} as an example to understand how to calculate conditional probability using a factor graph.
The variables $x_1, x_2, x_3 \in X$, and y $is$ the outcome.
$P(x_1,x_2,x_3)$ can be factored as a product of $f_1, f_2, f_3$. 
\begin{equation}
	P(x_1, x_2, x_3) = f_1(x_1), f_2(x_1,x_2), f_3(x_2, x_3)
	\label{eqn:6}
\end{equation}
Thus, the conditional probability of an event given the observed variables is calculated as: 
\begin{equation}
	P(y|x_1,x_2,x_3) = \prod_{f\in F} f(x)
	\label{eqn:7}
\end{equation}
Using sum-product algorithm, the marginal is defined as $P(x) = \sum_{x \in X} P(X)$. 
Thus, the conditional probability of an event given the observed variables is calculated as the following.
\begin{equation}
	P(y|x_1,x_2,x_3) = \sum f(x)
\end{equation}


\subsection{Extended Kalman Filter Implementation}
\label{app:ekf}
Extended Kalman Filter (EKF) is a version of Kalman Filter that is used to estimate the state dynamics of a non-linear system. 
EKF consists of two steps: estimation and correction. 
State estimation is done using a non-linear function of the system $f(.)$.
The following equation shows EKF's state estimation, where $x_t$ is the state estimated in the previous time step, $u_t$ is the actuator command. 

\begin{equation}
    \thinspace
    x_{t+1} = f(x_t, u_t)
\end{equation}

Equation~\ref{eqn:ekf} shows how EKF state correction is done using the available sensor measurements. 
$y_t$ is the sensor measurement, $h(\hat{x}_{t| t-1)}$ are EKF computed states in the previous time step, and $K$ is the Kalman Gain. Note that the Kalman Gain $K$ is calculated using Jacobian of transition and observation matrices and the sensor noise covariance matrix. 
 
\begin{equation}
    \thinspace
    \hat{x}_{t+1} = f(x_t, u_t) + K_t(y_t - h(\hat{x}_{t| t-1)})
    \label{eqn:ekf}
\end{equation}

We use non-linear system dynamics equations of RVs as models in our EKF estimator and for each RV we learn the model parameters (unknown coefficients) using system identification~\cite{matlab-si}. 
We collect a dataset of control actions and sensor measurements for each of our subject RVs. 
For example, we run missions capturing sensor readings and control signals to the rotors in various modes of operation of a drone - takeoff, loiter, auto, circle, and land. 

A quadcopter's state vector represents its position ($x, y, z$), velocity ($v_x, v_y, v_z$), angular orientation($\phi, \theta, \psi$), and angular speed ($\omega_\phi, \omega_\theta, \omega_\psi$). The following equations show the quadcopter dynamics~\cite{savior}. $U_t$ is the thrust force, $U_\phi, U_\theta, U_\psi$ are control commands for rotors (roll, pitch, yaw), $I_x, I_y, I_z$ are moments of inertia in x, y, z axis respectively, $g$ is gravity, and $m$ is the mass.

\begin{align*}
& \dot{x} = v_x \\
& \dot{y} = v_y \\
& \dot{z} = v_z \\
& \dot{v}_x = \frac{U_t}{m} \left( \cos \phi \sin \theta \cos \psi + \sin \phi \sin \psi \right) \\
& \dot{v}_y = \frac{U_t}{m} \left( \cos \phi \sin \theta \sin \psi - \sin \phi \cos \psi \right) \\
& \dot{v}_z = \frac{U_t}{m} \cos \phi \cos \theta - g \\
& \dot{\phi} = \omega_\phi \\
& \dot{\theta} = \omega_\theta \\
& \dot{\psi} = \omega_\psi \\
& \dot{\omega_\phi} = \frac{U_{\phi}}{I_{x}} + \dot{\theta}\dot{\psi}\left(\frac{I_{y}-I_{z}}{I_{x}}\right) \\
& \dot{\omega_\theta} = \frac{U_{\theta}}{I_{y}} + \dot{\phi}\dot{\psi}\left(\frac{I_{z}-I_{x}}{I_{y}}\right) \\
& \dot{\omega_\psi} = \frac{U_{\psi}}{I_{z}} + \dot{\phi}\dot{\theta}\left(\frac{I_{x}-I_{y}}{I_{z}}\right)
\end{align*}

The following equations show ground rover dynamics~\cite{model-rover}, where $\beta$ is the slip angle - measures sideways movement of wheels, $\delta$ is the steering angle, $l_f$ is the distance between center of mass and front axle, $l_r$ is the distance between center of mass and rear axle. 

\begin{align*}
& \beta = \tan^{-1}\left( \frac{l_r}{l_f + l_r} \tan(\delta) \right) \\
& \dot{x} = v \cos(\psi + \beta) \\
& \dot{y} = v \sin(\psi + \beta) \\
& \dot{\psi} = \frac{v}{l_r} \sin(\beta) \\
& \dot{v} = a
\end{align*}

\subsection{Algorithm}
\label{app:algorithm}
Algorithm~\ref{algo:recovery} shows \sysname's algorithm for attack diagnosis and recovery.
In the absence of attacks (based on the attack detector's response), \sysname records the physical states associated with all the on-board sensors (Lines 7-15) in a sliding window. 
Once the attack detector raises an alert, \sysname activates recovery mode (Line 17). 
First, it stops recording, and prepares the most recent safe $HS$ for recovery.
Then, it activates the diagnosis procedure (Line 21). 
The attack diagnosis module determines the sensors under attack, if any, using FGs (Line 28-34).
Once the targeted sensors are determined, \sysname uses the historic states for the compromised sensors and current state estimates from uncompromised sensors to reconstruct RV's states (Line 23), effectively representing the RV's states at the time when attack recovery was activated. 
This ensures that the inputs to the recovery technique remain untainted by SDA manipulations, enabling the recovery technique to derive safe control actions.
Once the attack subsides (based on the attack detector's response), the recovery mode is turned off.

\begin{algorithm}[!ht]
	\footnotesize
	\caption{Algorithm for Attack Diagnosis and Recovery}
	\label{algo:recovery}
	\begin{algorithmic}[1]
		\State $w \leftarrow$ historic state recording window
		\State $S \leftarrow$ historic physical states used for recovery
		\State $x_r \leftarrow$ physical states corresponding to targeted sensors
		\State $x_s \leftarrow$ physical states corresponding to uncompromised sensors
		\Procedure{AttackRecovery}{}
			\State $alert \leftarrow AttackDetector()$
			\While{!$mission\_end$}
				\If{$!alert$} \Comment{recording historic states}
				\State $recovery\_mode \leftarrow false$
					\If{$t_w < t_{w+1}$}
						\State $record \leftarrow true$
						\State $w_i[..] \leftarrow x(t)$
					\Else
						\State $delete~w_{i-1}$ 
						\State $t_w = t_{w+1}$
					\EndIf	
				\Else
					\State $recovery\_mode \leftarrow true$ \Comment{recovery activated}
					\State $record \leftarrow False$ 
					\State $HS = w_{i-1}$
					\State $targetedSensors \leftarrow Diagnosis()$
					\State $x_r \leftarrow estimateStates(HS[..])$
					\State $X_r[..] \leftarrow stateReconstruction(x_r, x_s)$	\Comment{state reconstruction}				
				\EndIf
			\EndWhile
		\EndProcedure
		\Procedure{Diagnosis}{}
			\While{!$mission\_end~\forall~sensors$}
				\State $e = |(e_t) - (e_{t-1})|$
				\State $s_t \leftarrow argmax P(s_t|e)$ \Comment{most probable outcome}
				\If{$s_t = 1$} 
					\State \textbf{return}~$malicious$ \Comment{targeted sensors}
				\EndIf
			\EndWhile
		\EndProcedure
	\end{algorithmic}
\end{algorithm}

\subsection{Mission Types}
\label{app:mission-types}
We run a diverse set of missions to derive the experimental parameters and for our experiments. Table~\ref{tab:mission-paths} shows the details.  

\begin{table}[h]
	\centering
	\footnotesize
	\caption{Mission paths used in evaluating \sysname. 
		S: Straight line, MW: Multiple waypoints, C: Circular, and Three different polygonal paths P1, P2, and P3}
	\begin{tabular}{l|l|l|l|l|l|l|l}
		\hline
		\textbf{\begin{tabular}[c]{@{}l@{}}Mission\\ Paths\end{tabular}}       & \textbf{S} & \textbf{MW} & \textbf{C} & \textbf{P1} & \textbf{P2} & \textbf{P3} & \textbf{Total} \\ \hline 
		\textbf{\begin{tabular}[c]{@{}l@{}}Number of \\ missions\end{tabular}} & 70                   & 70                   & 50                & 50                   & 50                   & 50                   & 340            \\ \hline
	\end{tabular}
	\label{tab:mission-paths}
\end{table}

\subsection{Metrics}
\label{app:metrics}
\noindent
\textbf{Normalized RMSD}
As minor attitude deviations are expected in RV missions even in the same trajectory due to variations in control signals and computations in RV's feedback control loop. 
To ensure that these minor deviations are accounted for in our recovery stability metric, we introduce a normalization technique for the RMSD values. 
This normalization approach transforms the RMSD values (calculation shown in Equation~\ref{eqn:rmsd}) into a standardized range between 0 and 1 as follows:  
 
\begin{equation}
    Normaized~RMSD = \frac{RMSD-min~RMSD}{max~RMSD-min~RMSD}
    \label{eqn:nrmsd}
\end{equation}

\medskip
Where $min~RMSD$ and $max~RMSD$ are minimum and maximum RMSD values among recovery activated missions. 

\smallskip
\noindent
\textbf{Normalized Mission Delay}
As minor variations in mission completion times can occur even on the same trajectory, we normalize the mission delay using a baseline mission completion time $T_{baseline}$ as the reference point for mission completion time for an RV on a given trajectory. 
$T_{baseline}$ is derived using min-max approach as shown below. Where $T_{baseline}$ is the average of the minimum and maximum mission completion times in attack-free missions.  

\begin{equation}
    T_{baseline} = \frac{T_{min} + T_{max}}{2}
\end{equation}

\end{document}